\documentclass{bmvc2k}

\title{Semi-Supervised Domain Generalization for Object Detection via Language-Guided Feature Alignment}
\addauthor{Sina Malakouti}{sem238@pitt.edu}{1}
\addauthor{Adriana Kovashka}{kovashka@cs.pitt.edu}{1}

\addinstitution{
University of Pittsburgh, \\
Pittsburgh, PA, USA
}

\runninghead{S.M \& A.K }{Semi-Sup. Domain Gen. for Obj. Det. via Lang.-Guided Feat. Align}

\usepackage{spreadtab}
\usepackage{siunitx}
\usepackage{makecell}
\usepackage{booktabs, multirow} 
\usepackage{soul}
\usepackage{changepage,threeparttable} 
\usepackage{longtable}
\usepackage{color}
\usepackage{colortbl}
\usepackage{amssymb}
\usepackage{tablefootnote}
\usepackage{xfp}
\usepackage{enumitem}

\begin{document}

\maketitle

\begin{abstract}
Existing domain adaptation (DA) and generalization (DG) methods in object detection enforce feature alignment in the visual space but face challenges like object appearance variability and scene complexity, which make it difficult to distinguish between objects and achieve accurate detection. In this paper, we are the first to address the problem of semi-supervised domain generalization by exploring vision-language pre-training and enforcing feature alignment through the language space. We employ a novel Cross-Domain Descriptive Multi-Scale Learning (CDDMSL) aiming to maximize the agreement between descriptions of an image presented with different domain-specific characteristics in the embedding space. CDDMSL significantly outperforms existing methods, achieving $11.7\%$ and $7.5\%$ improvement in DG and DA settings, respectively. Comprehensive analysis and ablation studies confirm the effectiveness of our method, positioning CDDMSL as a promising approach for domain generalization in object detection tasks. Our code is available at \href{https://github.com/sinamalakouti/CDDMSL}{https://github.com/sinamalakouti/CDDMSL}.

\end{abstract}



\section{Introduction}
\label{sec:intro}
The recent success of fully-supervised object detectors (FSOD) heavily relies on the assumption that training data follows the same distribution as test data, which often fails in real-world applications. Domain adaptation (DA) addresses distribution alignment between a source and a particular target domain but struggles to generalize to domains unseen during training, limiting its real-world applicability 
\cite{zhou2021domain,lin2021domain,wu2022single}. On the other hand, domain generalization (DG) aims to learn a universal model that can generalize to any unseen domain. While DG has been studied in object recognition, DG in object detection (DGOD) is still heavily understudied. Pioneering work \cite{lin2021domain} utilizes multiple fully annotated source domains to learn domain-invariant features. However, obtaining multiple fully annotated labeled datasets 
is a formidable challenge in object detection. 

Fortunately, obtaining unlabeled data is less intricate, and a large volume is available. To tackle the issue above, we formulate DGOD as a semi-supervised learning (SSL) paradigm, using one fully annotated and an additional unlabeled source domain. While semi-supervised domain generalization (SSDG) \cite{zhou2021semi,zhou2021mixstyle,liu2021semi,shu2021open} has been studied in image recognition, to the best of our knowledge, this is the first work addressing SSDG in object detection.

\begin{figure}[!tp]
\begin{subfigure}{0.49\textwidth}
\centering
\scriptsize
    \includegraphics[width=0.95\textwidth]{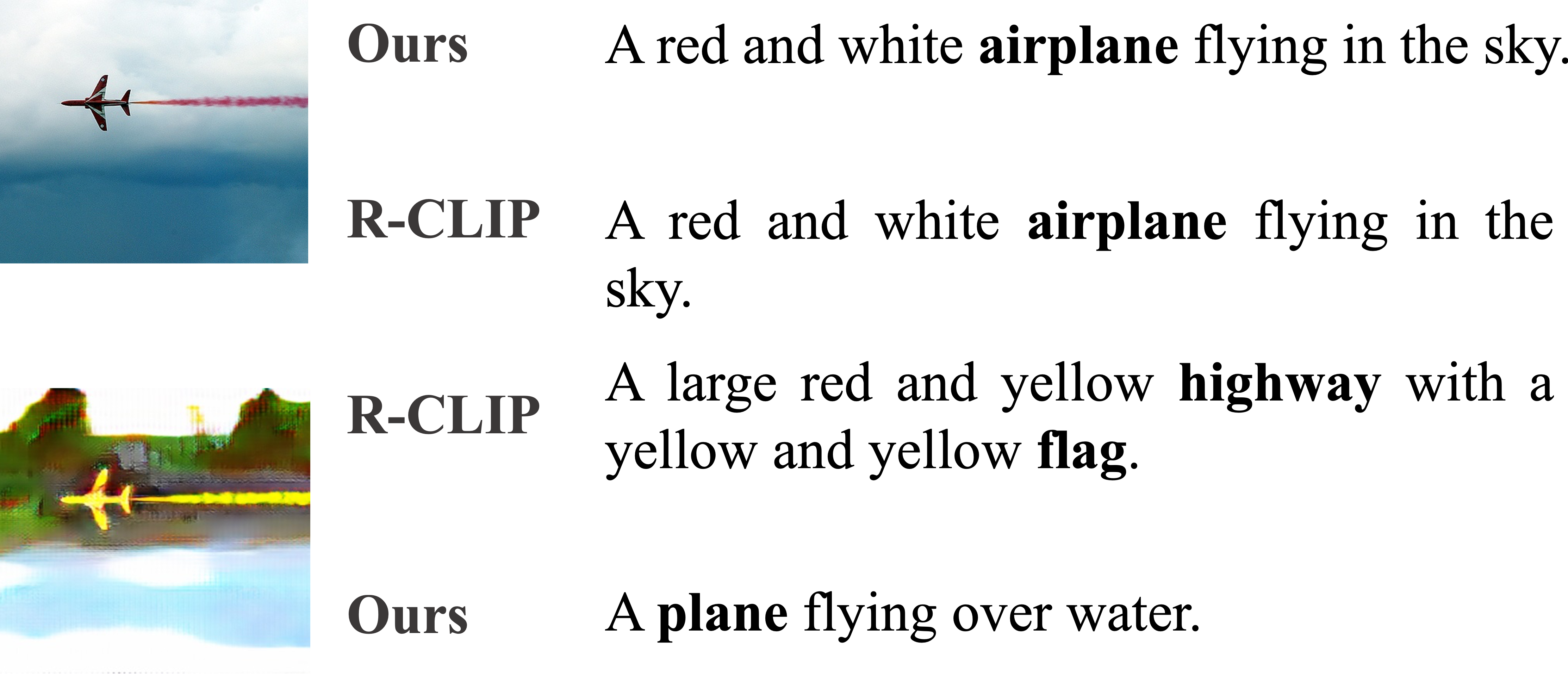}
    \caption{Motivation}
    \label{fig:B}
\end{subfigure}
\begin{subfigure}{0.5\textwidth}
\centering
    \includegraphics[width=1.0\textwidth]{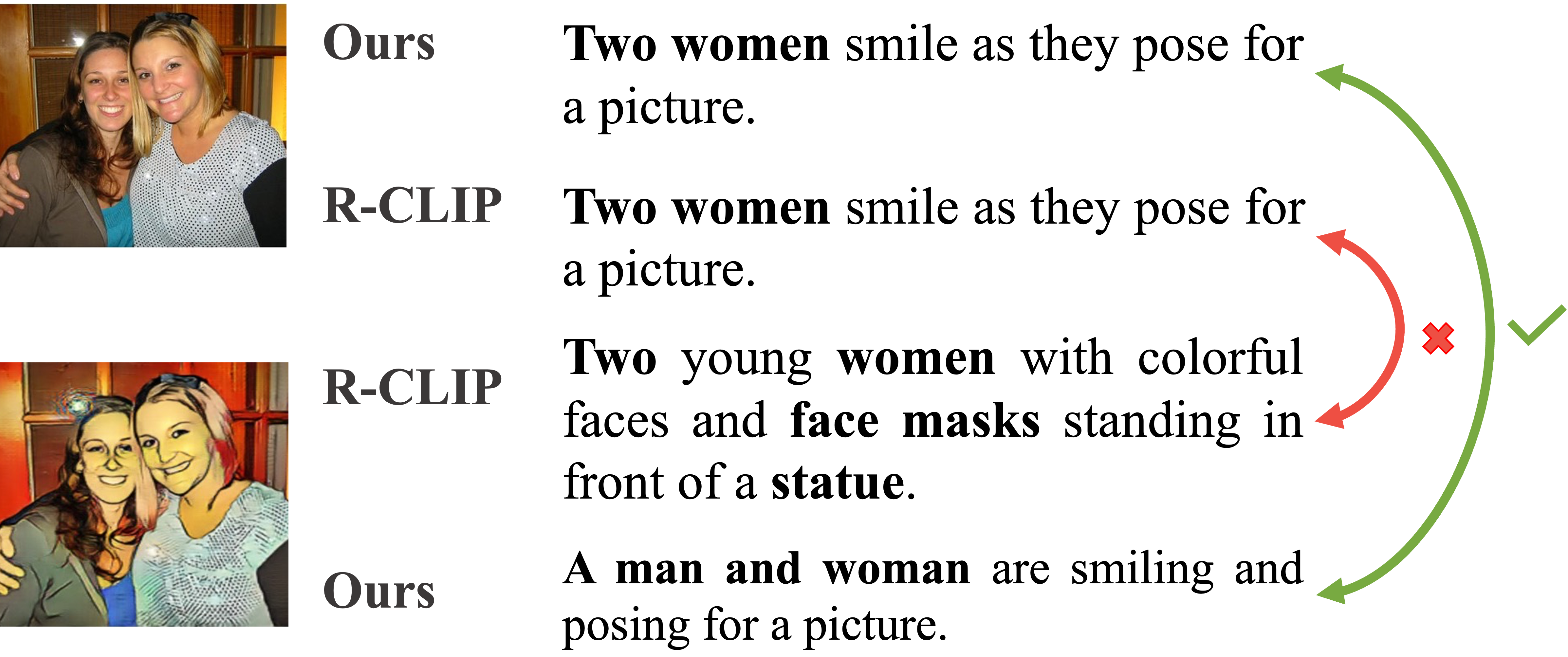}
    \caption{Consistency Objective}
    \label{fig:A}
\end{subfigure}
\caption{(a) Demonstrates that an image-captioning model, when trained with RegionCLIP, fails to produce descriptions across different styles. In contrast, CDDMSL effectively maintains overall image semantics across domains. (b) Our consistency objective is designed to preserve the overarching image semantics regardless of differing styles (green) while discouraging unrelated descriptions (red).}
\label{fig:both}
\end{figure}

Generally, existing DA and DG for object detection works learn invariant representations or disentangle domain-specific representations by enforcing their objectives in the visual space \cite{oza2023unsupervised,zheng2020cross,hsu2020every,li2022cross,saito2019strong,oza2023unsupervised,deng2021unbiased,wu2022single}. However, disentangling domain-specific and domain-invariant features in object detection is extremely difficult due to inherent challenges such as variability in object appearance (e.g., size, shape), and scene complexity, which can make it difficult to distinguish between objects and accurately detect them. We believe high-level information, such as object relations, visible object characteristics, or actions being performed, can help perform better under extreme domain shifts. Extracting such contextual information is infeasible using visual features solely, but textual descriptions offer a semantically denser signal \cite{desai2021virtex} that can help the model to focus on the underlying semantic information of the data, resulting in better localization and recognition of objects in complex scenes. Also, natural language descriptions provide more information than simple object labels; the latter may be ambiguous or insufficient. Thus, we enforce semantic consistency in the textual descriptions of images to ensure that useful information is preserved across domains. 

We believe that by incorporating natural language captions in object detection models, the models can learn to associate object categories with their names and other semantic attributes, such as actions, relations, and context. This can help the models better generalize to new or unseen domains where objects' visual appearance may differ from what was observed in the training data. Note that previous works have shown the benefit of using image captions in object detection tasks, such as improving the performance of weakly supervised object detectors and generalizing over novel categories in open-vocabulary object detection tasks \cite{gupta2017aligned,Ye_2019_ICCV,zhong2022regionclip,zareian2021open,Ye_2019_ICCV,Datta_2019_ICCV,9811398}. However, to the best of our knowledge, this is the first work to explore the benefits of incorporating image descriptions and vision-language pre-training in the context of generalizability to unseen domains for object detection. 

In this paper, we first make three observations: (i) vision-language models, such as CLIP \cite{radford2021learning}, have shown promise in improving generalizability in image recognition tasks, (ii) captions contain rich semantic information that can be helpful in both object localization and detection, and (iii) as illustrated in Fig.~\ref{fig:B}, domain shift impacts the ability of image-captioning models to produce semantically consistent descriptions for the same image with different domain-specific characteristics (i.e., style). 
Based on these observations, we present a simple yet effective novel approach called Cross-Domain Descriptive Multi-Scale Learning (CDDMSL) for object detection, which enforces learning domain-invariant features on the visual encoder through the language space. In particular, we first propose to initialize the object detector's backbone with RegionCLIP \cite{zhong2022regionclip} weights, a state-of-the-art vision-language model in object detection, and adopt its text classifier to facilitate supervised object detection on the labeled data. Second, we exploit unlabeled data to learn domain-invariant features while preserving the semantics. To this end, we utilize the vision-to-language module of an image-captioning model to obtain descriptive features by projecting image features to the language space. To ensure semantically consistent descriptions, we apply a contrastive-based consistency objective over generated descriptive features of an image and its stylized version synthesized by a style transfer model (shown in Fig.~\ref{fig:A}).
We enforce the contrastive objective in the embedding space, which is crucial for learning semantically consistent features, as demonstrated in Sec.~\ref{sec:ablation}. We extend the proposed method to remedy existing domain bias at both the image level and instance level \cite{lin2021domain,zhong2022regionclip}. Note our proposed approach only impacts the training and does not impose any computational overhead on the inference time. 

Our experiments demonstrate that the proposed CDDMSL approach significantly outperforms existing DG and DA methods on two object detection benchmarks. In particular, CDDMSL achieves $11.7\%$ ($28\%$) and $7.5\%$ ($20.8\%$) improvement over the RegionCLIP (Faster-RCNN) in DG and DA settings, respectively, and significantly outperforms the state-of-the-art methods. Moreover, we provide comprehensive analysis and ablation studies, highlighting the benefit of enforcing generalizability through language space and the effectiveness of each component of the proposed method. Our findings indicate that CDDMSL is a promising approach for addressing the domain generalization challenge in object detection, making it a valuable contribution to the field.


\section{Related Works}
\label{sec:Relatedwork}

\textbf{Domain Adaptation for Object Detection.} A common DA approach is to align the feature space by adversarial learning or directly minimizing the distance between the feature distributions in the visual space \cite{oza2023unsupervised,rodriguez2019domain,hsu2020every,chen2018domain,zheng2020cross}. Many works utilize self-training and highly confident predictions on the target (i.e., pseudo-labels) to progressively improve the model's performance on the target \cite{oza2023unsupervised,li2022cross,deng2021unbiased,roychowdhury2019selftrain}. For instance, \cite{li2022cross, deng2021unbiased} leverage a Mean Teacher framework \cite{NIPS2017_68053af2} to produce the pseudo-labels by the teacher model and guide the student to perform well on the target. MTOR \cite{cai2019exploring} performs Mean Teacher to explore object relation in region-level, inter-graph, and intragraph consistency. Another work utilizes an unpaired image-to-image translation model to map the source data to target-like images to reduce distribution shift \cite{oza2023unsupervised,rodriguez2019domain,inoue2018cross,8822427,chen2020htcn}. 
However, 
these methods require the pre-collection of target data and fail when the target domain is unknown. 

\textbf{Domain Generalization in Object Detection.} Unlike DA, there are not many works addressing domain generalization for object detection (DGOD). Initially, DIDN \cite{lin2021domain} relied on fully-annotated source domains and adversarial learning to learn domain-invariant features and disentangle domain-specific ones with the help of domain-specific encoders. However, collecting multiple fully annotated source domains is cumbersome and time-consuming. To alleviate this issue, Single-DGOD \cite{wu2022single} extended single-domain generalization (SDG) for object detection and employed self-distillation to disentangle domain-specific features. However, SDG methods rely on the quality of the labeled data and data augmentation techniques, reducing their ability to generalize over a wider range of domains. Unlike Single-DGOD, we use unlabeled data along with only one labeled domain, i.e., semi-supervised domain generalization (SSDG). Pseudo-labeling and data augmentation are some common SSDG approaches. StyleMatch \cite{zhou2021semi} propagates pseudo labels from the labeled domain to the unlabeled domain, and the style transfer model is used for data augmentation. MixStyle \cite{zhou2021mixstyle} inserted a plug-and-play augmentation module in convolutional layers to mix feature statistics between labeled and pseudo-labeled instances. However, the quality of pseudo-labels substantially degrades on out-of-domain data \cite{li2022cross,deng2021unbiased} due to the existing distribution shift and challenges induced by producing pseudo-labeled bounding boxes \cite{li2022cross,deng2021unbiased}. Unlike existing DA and DG methods, which learn domain-invariant features using objectives in the visual space, we propose to use text/descriptions. 

\textbf{Vision-Language Models.}
Various works have studied the benefit of learning image-text joint representations in various computer vision tasks \cite{gupta2017aligned, radford2021learning, dou2022coarse,li2020oscar,lu2019vilbert,jia2021scaling}. VirTex \cite{desai2021virtex} pre-trained a visual encoder using image-caption pairs, which are then transferred for different downstream tasks. CLIP \cite{radford2021learning} and ALIGN \cite{jia2021scaling} learn robust and rich visual features by performing cross-modal contrastive loss on large amounts of image-text pairs for open-vocabulary image classification. Recently, several studies used image-text pairs for open-vocabulary object detection. \cite{bansal2018zero,zareian2021open} learned joint embedding by matching region features to word/text embeddings. ViLD \cite{gu2021open} learned visual features by distilling from pre-trained CLIP encoders. RegionCLIP \cite{zhong2022regionclip} and GLIP \cite{li2022grounded} use region-text matching and phrase-grounding, respectively, to pre-train the visual encoder for OVD. 
In contrast to these works, we study the effectiveness of language for generalizing over domain shift. 

\section{Approach}
\label{section:overview}
Let $\mathcal{X}$ and $\mathcal{Y}$ represent input space and label space, respectively. A domain can be formally described as a joint distribution $P_{XY}$ sampled from $\mathcal{X} \times \mathcal{Y}$, where $\mathcal{X} \times \mathcal{Y}$ denotes the set of all probability distributions. Unlike conventional DG settings, in SSDG, we only have access to one fully labeled and one or more unlabeled source domains, defined as $\mathcal{S} = \{S_k\}_{k=1}^K$, where $S_1 = S_{\mathcal{L}} = {(x_i ^\mathcal{L}, y_i^\mathcal{L}})_{i=1}^{N_\mathcal{L}}$ is a labeled source of size ${N_\mathcal{L}}$, $\{S_k\}_{k=2}^K = \{S_{\mathcal{U}_k}\}_{k=2}^K$ is the $k$-th unlabeled source of size $N_{\mathcal{U}_k}$, $K$ is the number of source domains, and $\mathbf{y}_i^\mathcal{L} = (\mathbf{b}_i^\mathcal{L}, c_i^\mathcal{L})$ denotes the corresponding labels with bounding-box coordinates $\mathbf{b}$ and their associated categories $c$. Each source domain $S_{k}$ is associated with a joint distribution $P_{XY}^k$ and $P_{XY}^k \neq P_{XY}^{k^\prime}$ for $k \neq k^{\prime}$. In this work, we assume that the $\mathbf{y}_j\in \mathcal{Y}$ share the same set of classes across all domains. The ultimate goal is to utilize the information from the source domains to develop an object detection model capable of generalizing effectively to an unseen target domain $\mathcal{T} = \{\mathbf{x}_j^T\}_{j=1}^{N_T}$ with $N_T$ examples, which is drawn from an unknown distribution $P_{XY}^{T}$. 

\subsection{Cross-Domain Descriptive Multi-Scale Learning}
\label{sec:CDDMSL}
The primary motivation is to leverage the semantically rich information in image descriptions generated by image-captioning models to learn a robust object detector model. We enforce the backbone to learn domain-invariant features, enabling a caption generator to produce consistent descriptions for images with similar semantics but different domain-specific characteristics. This ensures that the model preserves the essential information across different styles and enhances the model's generalization capabilities across various domains. To this end, we propose a novel contrastive-based consistency to maximize the agreement between descriptions of an image across domains in the embedding space.

\begin{figure}
\centering
\includegraphics[width=1\textwidth]
{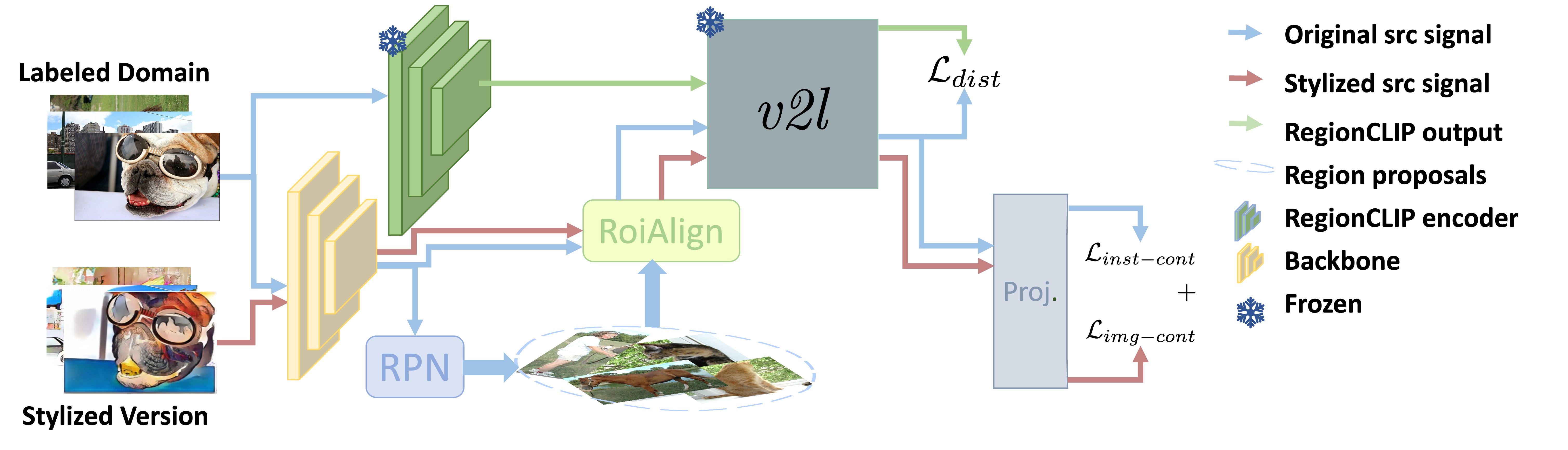}
\caption{\textbf{Overview of CDDMSL}. The backbone processes the image and its stylized version to produce image features. Region proposals are created for the labeled domain, and RoiAlign extracts region features from the original and stylized versions. The $\textit{v2l}$ module produces region descriptive features, projected to a lower dimension by $g(\cdot)$. 
The backbone and projection heads are trained to maximize agreement between descriptive features across domains using a contrastive loss $\mathcal{L}_{img-inst}$. 
Image-level training $\mathcal{L}_{img-cont}$ follows a similar process, with RoiAlign directly passing input features to $\textit{v2l}$. 
Finally, a knowledge distillation regularization loss $\mathcal{L}_{dist}$ is applied to the descriptive features of the labeled domain based on the backbone and RegionClip encoder output.}
\label{fig:model}
\end{figure}

As illustrated in Fig.~\ref{fig:model}, we use a vision-to-language (\textit{v2l}) module to obtain descriptive features by projecting image features to the language space. We propose to employ the mapping network of a pre-trained image-captioning model, which stays frozen throughout training. Specifically, we pre-train the ClipCap model \cite{mokady2021clipcap} with the object detector backbone $\mathbf{B}$ as its vision encoder and extract its mapping network. We apply \textit{v2l} to the image representation to obtain the desired descriptive features in the language space: $z^{\ell} = \textit{v2l}(\mathbf{B}(x))$. 
We define the proposed consistency learning approach in the following text. 

\textbf{Instance-level Descriptive Consistency Learning.} Consider a large labeled dataset $S_{\mathcal{L}}$ and a set of smaller unlabeled domains $S_{\mathcal{U}_k}$ (i.e., $N_{\mathcal{L}} >> N_{\mathcal{U}}$). We use the smaller domains to create a larger auxiliary domain $S_{\Tilde{\mathcal{U}}}$ of size $N_{\Tilde{\mathcal{U}}}$ = $(K-1) \cdot N_{\mathcal{L}}+\sum_{k=2}^{K} N_{{\mathcal{U}_k}}$ by synthesizing $S_{\mathcal{L}}$ to K-1 unlabeled source domains. Particularly, we use CycleGAN \cite{zhu2017unpaired} 
to stylize the images between the labeled and unlabeled source domains.
For simplicity, in the rest of the paper, we assume that images are translated from a labeled domain to an unlabeled (i.e., $N_{\Tilde{\mathcal{U}}} = N_{\mathcal{L}}$ and $K=2$), but 
the unlabeled domain(s) can also be transferred to labeled in order to enforce consistency. Let us define $\{(x_i, \Tilde{x_i}) | x_i \in S_{\mathcal{L}}, \Tilde{x}_i \in S_{\mathcal{\Tilde{U}}}\}$ where $x_i$ and $\Tilde{x}_i$ represent the same scene but belong to $S_{\mathcal{L}}$ and $S_{\mathcal{\Tilde{U}}}$, respectively. 
RPN \cite{ren2015faster} generates region proposal $r_{i,j}$ for image $x_i$, which is then used to extract region features $v_{i,j}$ and $\Tilde{v}_{i,j}$ by applying \text{RoIAlign} \cite{he2017mask} on ${z}_{i}^{v} = \mathbf{B}(x_i)$ and $\Tilde{z}_{i}^{v} = \mathbf{B}(\Tilde{x_i})$, respectively. Finally, descriptive region features can be computed by applying the $\textit{v2l}$ on each region feature, denoted as $z_{i,j}^{\ell} = \textit{v2l} (v_{i,j})$ and $\Tilde{z}_{i,j}^{\ell} = \textit{v2l} (\Tilde{v}_{i,j})$. For brevity, let us define all region descriptive feature pairs of all images in the batch as $Z^{\ell} =$ $\{ (z_{i}^{\ell},\Tilde{z}_{j}^{\ell})\}$, where $i \in \{1, \dots, N\}$, $j \in \{1, \dots, N\}$, $(z_{i}^{\ell},\Tilde{z}_{i}^{\ell})$ is a positive pair, $(z_{i}^{\ell},\Tilde{z}_{j}^{\ell})_{i\neq j}$ is a negative pair, and $N$ is the total number of the proposals or region descriptive features in the batch. The instance-level contrastive loss can be defined:
\begin{gather}
  \mathcal{L}_{inst-cont} = \frac{1}{N} \sum_i -\log\left(\frac{\exp(\mathrm{s}_{i,i} / \tau)}{ exp(\mathrm{s}_{i,i}/ \tau)+\sum_{k} \exp(\mathrm{s}_{i,k} / \tau)}\right) ;  \, \, \, k \neq i, \, \, h_i = g(z^{\ell}_i) \label{eq:inst-cont} \\
  \mathrm{s}_{i,j} =  \mathrm{s} (h_i,\Tilde{h}_j)= \frac{h_{i}^{\top} \cdot \Tilde{h}_j}{\vert \vert h_i \vert \vert  \cdot \vert \vert \Tilde{h}_j \vert \vert } \label{eq:cosine-sim}
\end{gather}
where $g(\cdot)$ is a projection network to project region descriptive features to a lower dimension, $s_{i,j}$ = is a cosine similarity score, $\mathrm{s}_{i,i} = \mathrm{s} (h_i, \Tilde{h}_i)$ is a positive pair, $\mathrm{s}_{i,k} = \mathrm{s} (h_i, \Tilde{h}_k)$ is a negative pair, and $\tau$ is temperature parameter. One key aspect of our method is the choice of applying the contrastive loss in the embedding space (i.e., the output of the \textit{v2l} layer) rather than directly in the language space (i.e., on the tokens). This choice has several advantages. The embedding space provides a continuous and compact representation of semantic content, enabling meaningful similarity metrics between images. By focusing on learning semantically consistent features in the embedding space, the model is less sensitive to specific phrasing or word choice. Additionally, optimizing in the embedding space leads to a more efficient optimization process without needing to backpropagate gradients through the entire language generation process, which is non-differentiable.

\textbf{Image-level Descriptive Consistency Learning.} The proposed instance-level contrastive loss can naturally extend to an image-level contrastive loss $\mathcal{L}_{img-cont}$ by simply replacing the region features with image features in Eq.~\ref{eq:inst-cont}. We thus encourage the generation of semantically consistent descriptive features in different domains by minimizing the contrastive loss at both image-level and instance-level. This removes the domain bias in the backbone $\mathbf{B}$ and results in a model capable of learning domain-invariant visual features at multiple scales, leading to better generalization performance on unseen target domains.

\subsection{Regularization via Knowledge Distillation}
\label{sec:KD}
Merely applying the contrastive loss may lead to converging to a trivial solution, generating consistent but meaningless descriptions. 
We propose a knowledge distillation-based (KD) regularization by exploiting that the image-captioning model produces meaningful descriptions, utilizing pre-trained RegionCLIP as its image-encoder. Concretely, for a given image $x_i \in \mathcal{S}_L$, the regularization loss is defined as:
\begin{equation}
\mathcal{L}_{dist} = \frac{1}{N_{\mathcal{L}}}\sum_{i=1}^{N_{\mathcal{L}}}\mathbf{d}(\textit{v2l}(z_i), \textit{v2l}(z_i^R)); \, \, \,  \ z_i^R = F_{R-CLIP}(x_i)
\end{equation}
$z_i$ and $z_i^R$ represent the model's backbone (yellow encoder in Fig.~\ref{fig:model}) and the RegionCLIP ($F_{R-CLIP}$) output, respectively. $\mathbf{d(\cdot,\cdot)}$ is a distance function. In this work, the Manhattan distance metric is used. It is crucial to note that the RegionCLIP features remain frozen throughout the training.
\subsection{Object Detector Training}
\label{sec:supervised}
We use the labeled source domain $S_{\mathcal{L}} = {(x_i^\mathcal{L}, y_i^\mathcal{L}})_{i=1}^{N_\mathcal{L}}$ to train a Faster-RCNN \cite{ren2015faster}. Unlike Faster-RCNN, we adopt a text-based classifier in the detection head. 
Following RegionCLIP, we create prompts for each object category by filling it into prompt templates (e.g., "A photo of a car," "A painting of a bus," etc.) 
and encode the prompts using the pre-trained CLIP language encoder. Finally, cross-entropy with focal scaling \cite{pereyra2017regularizing} is employed as the classification loss $\mathcal{L}_{cls}$.
We use fixed all-zero embedding for the background category and apply a predefined weight to background regions following \cite{zhong2022regionclip,zareian2021open}.
The detector is then trained with a $\mathcal{L}_{det}$ loss which combines $\mathcal{L}_{cls}$ with standard RPN and regression loss as in \cite{ren2015faster}: 
$    \mathcal{L}_{det} = \mathcal{L}_{cls} + \mathcal{L}_{reg} + \mathcal{L}_{RPN}$.
Lastly, training consists of two major stages. During the burn-up stage, the object detector is warmed-up in a supervised manner, whereas in joint training, it is jointly learned by supervised and contrastive objectives: $\mathcal{L}_{tot} = \mathcal{L}_{det} + \mathcal{L}_{inst-contr} + \mathcal{L}_{img-contr} + \omega \cdot \mathcal{L}_{dist}$, where $\omega$ is a regularization parameter. 
\section{Experiments}
\label{sec:experiments} 
\textbf{Real-to-artistic generalization.} We conduct our experiments based on 4 
datasets: Pascal-VOC (VOC) \cite{pascal-voc-2012,pascal-voc-2007}, Clipart1k \cite{inoue2018cross}, Watercolor2k \cite{inoue2018cross}, Comic2k \cite{inoue2018cross} to assess generalizability on artistic and style domain shifts. VOC contains 20 categories of common objects from 16,551 natural images, Clipart consists of 20 categories of 3,165 objects from 1000 instances, and Watercolor and Comic contain 6 categories of 3,315 and 6,389 objects from 2,000 images, respectively. Experiments are conducted in three settings: (1) VOC and Clipart, (2) VOC and Watercolor, and (3) VOC and Comic are used as the source domains, in which only VOC is considered a labeled domain. In each setting, the remaining datasets are considered unseen targets 
to assess the model's generalizability across multiple domains (Table~\ref{tab:DG-voc}). We extend our model to DA and show the results in Table~\ref{tab:DA-voc}.

\textbf{Adverse weather adaptation.} 
We experiment on Cityscapes \cite{Cordts2016Cityscapes} (City), Foggy-Cityscapes \cite{Cordts2016Cityscapes} (Foggy), and BDD100k \cite{bdd100k} (Bdd) to evaluate generalizability across different time and weather conditions. City features 3,025 urban images from 50 cities, Foggy simulates fog on the City, and Bdd comprises 10,000 images of various weather conditions and times. All datasets contain 8 categories, but we ignore the ``train'' category on Bdd due to its low instance count in the validation set \cite{lin2021domain}. As Foggy is a simulated based on City, we use them as the source domains without the need for style transfer (City labeled and Foggy unlabeled), while Bdd is an unseen domain in the domain generalization task summarized in Table~\ref{tab:DG-city}. Note we do not use the labels on the Foggy data throughout training. Following DIDN \cite{lin2021domain}, since most previous works are on weather adaptation, we perform adaptation from City to Foggy and extensively compare against baselines and related works as shown in Table~\ref{tab:uda-c2fc}.

\textbf{Baselines.}
\label{sec:baselines}
Faster-RCNN \cite{ren2015faster} (F-RCNN) with ImageNet \cite{deng2009imagenet} and RegionCLIP (R-CLIP) are labeled source-only baselines, trained on VOC for real-to-artistic generalization/adaptation (Table~\ref{tab:DG-voc} / Table \ref{tab:DA-voc}) and City for adverse weather generalization/adaptation (Table~\ref{tab:DG-city} / Table~\ref{tab:uda-c2fc}). For a fair comparison, we only compare against models using ResNet50 backbone as RegionCLIP \cite{zhong2022regionclip} does not have ResNet101 pre-trained weights available. For real-to-artistic generalization, we further train Adaptive-MT
\cite{li2022cross}, a state-of-the-art (sota) DA with ResNet50 backbone. We define \textit{Direct Visual Alignment} (DVA) and \textit{Caption Pseudo Labeling} (Caption-PL) as additional baselines in the ablation (Table~\ref{tab:DG-voc} and Table~\ref{tab:DA-voc} for DG and DA in real-to-artistic, respectively). DVA applies the descriptive consistency loss in the visual space without using $v2l$. Caption-PL uses pseudo labels obtained by applying ClipCap on the original image to compute image-captioning loss over tokens for the stylized image. IRG \cite{vs2022instance} is only reported for the (VOC and Clipart) setting 
as performance/pre-trained weights were reported only in this setting. 

\textbf{Implementation details.} PyTorch \cite{NEURIPS2019_9015} and Detectron2 \cite{wu2019detectron2} are used for development. We adopt RegionCLIP's ResNet50 \cite{he2016deep} as the object detector backbone. Following ClipCAP\footnote{\url{https://github.com/rmokady/CLIP_prefix_caption}}, $v2l$ comprises 8 multi-head self-attention layers, each with 8 heads. ClipCAP is pre-trained on COCO-captions \cite{chen2015microsoft,lin2014microsoft} using frozen RegionCLIP as the image encoder and GPT-2 as the language decoder, with only the mapper ($v2l$) being updated. Projection layer $g$ (after $\textit{v2l}$ in Fig.~\ref{fig:model}) has output dimension of 256. For all experiments, we use SGD with 0.002 initial lr, a linear scheduler, and a batch size of 8, except 12 for the RegionCLIP baseline. Experiments use 4 NVIDIA Quadro RTX 5000 with 10k burn-up and 20k joint-training iterations. The burn-up stage follows RegionCLIP \cite{zhong2022regionclip}, and SSDG experiments continue from this checkpoint. $\omega$ in $\mathcal{L}_{tot}$ is 1. We evaluate mAP with a 0.5 threshold. For realistic evaluation, following \cite{cct,oliver2018realistic}, we avoided intensive hyperparameter search and manually chose reasonable parameters that resulted in stable training. However, a comprehensive search is expected to improve the performance.
\subsection{Results}

\begin{table}[!tp]
\scriptsize
\centering
  \captionsetup{skip=0pt, position=above} 
 \caption{\textbf{Real-to-artistic generalizations}. Numbers in parentheses show the difference from R-CLIP. Max $\uparrow$ shows maximum improvement over two target domains compared to F-RCNN. We report mAP ($\%$). $\dagger$/$\ddagger$ denote DA methods/labeled source-only methods.
 }
  \label{tab:DG-voc}
  \begin{tabular}{@{\hspace{3pt}}llllllll@{\hspace{1pt}}} 
\hline
    \multirow{2}{*}{{Method}}&
    \multicolumn{2}{l}{{VOC\&Clip \textrightarrow Water,Com}}  & 
    \multicolumn{2}{l}{{VOC\&Water\textrightarrow Clip,Com}} &
    \multicolumn{2}{l}{{VOC\&Com\textrightarrow Clip,Water}}&
    \multirow{2}{*}{Max $\uparrow$ }\\  
\cmidrule(lr){2-3}\cmidrule(lr){4-5}\cmidrule{6-7}
& Watercolor&Comic&Clipart&Comic&Clipart&Watercolor&\\
\hline
\multicolumn{1}{l|}{F-RCNN$\ddagger$ \cite{ren2015faster}} &41.2&17.9&24.1&17.9&24.1&41.2&\multicolumn{1}{|l}{-}\\
\multicolumn{1}{l|}{R-CLIP$\ddagger$ \cite{zhong2022regionclip}}  &44.7&34.2&33.9&34.2&33.9&44.7&\multicolumn{1}{|@{\hspace{3pt}}l@{\hspace{3pt}}}{16.3/16.3/9.8}\\
\hline
\multicolumn{1}{l|}{Adaptive MT$\dagger$ \cite{li2022cross}} &40.6 \tiny{\textcolor{red}{(-4.1)}}&22.2 \tiny{\textcolor{red}{(-12.0)}}&29.0 \tiny{\textcolor{red}{(-4.9)}}&24.3 \tiny{\textcolor{red}{(-9.9)}}&25.7 \tiny{\textcolor{red}{(-8.2)}}&42.3 \tiny{\textcolor{red}{(-2.4)}}&\multicolumn{1}{|@{\hspace{3pt}}l@{\hspace{0pt}}}{4.3/6.4/1.6}\\
\multicolumn{1}{l|}{IRG$\dagger$ \cite{vs2022instance}} &48.1 \tiny{\textcolor{blue}{(+3.4)}}&25.9 \tiny{\textcolor{red}{(-8.3)}}&-&-&-&-&\multicolumn{1}{|@{\hspace{3pt}}l@{\hspace{3pt}}}{8.0/-/-}\\
\hline
\multicolumn{1}{l|}{DVA} &45.6 \tiny{\textcolor{blue}{(+0.9)}}&38.1 \tiny{\textcolor{blue}{(+3.9)}}&
32.6 \tiny{\textcolor{red}{(-1.3)}} &34.2 \tiny{\textcolor{blue}{(+0.0)}}&35.9 \tiny{\textcolor{blue}{(+2.0)}}&45.9 \tiny{\textcolor{blue}{(+1.2)}}&\multicolumn{1}{|@{\hspace{3pt}}l@{\hspace{0pt}}}{20.2/16.3/11.8}\\
\multicolumn{1}{l|}{Caption-PL} &45.0 \tiny{\textcolor{blue}{(+0.3)}}&36.4 \tiny{\textcolor{blue}{(+2.2)}}&30.1 \tiny{\textcolor{red}{(-3.8)}}&30.3 \tiny{\textcolor{red}{(-3.9)}}&34.7 \tiny{\textcolor{blue}{(+0.8)}}&42.1 \tiny{\textcolor{red}{(-2.6)}}&\multicolumn{1}{|@{\hspace{3pt}}l@{\hspace{0pt}}}{18.5/12.4/10.6}\\

\multicolumn{1}{l|}{\textbf{Ours}}  &\textbf{49.8} \tiny{\textcolor{blue}{(+5.1)}}&\textbf{45.9} \tiny{\textcolor{blue}{(+11.7)}}&\textbf{38.7} \tiny{\textcolor{blue}{(+4.8)}}&\textbf{43.5} \tiny{\textcolor{blue}{(+9.3)}}&\textbf{39.8} \tiny{\textcolor{blue}{(+5.9)}}&\textbf{49.4} \tiny{\textcolor{blue}{(+4.7)}}&\multicolumn{1}{|@{\hspace{3pt}}l@{\hspace{3pt}}}{\textbf{28.0}/\textbf{25.6}/\textbf{15.7}}\\
\hline
\end{tabular}
\end{table}

\begin{table}[!tp]\centering

\scriptsize
\captionsetup{skip=0pt, position=above} 

\caption{\textbf{Adverse weather generalization.} \emph{City, Foggy \textrightarrow \, Bdd}  in $\%$}
\label{tab:DG-city}
\begin{tabular}{l@{\hspace{3pt}}|lllllll@{\hspace{2pt}}|@{\hspace{3pt}}l@{\hspace{3pt}}}
\hline
{Method}  &{prsn} &{rider} &{car} &{truck} &{bus} &{motor} 
&{bike} & {mAP} \\
\hline
F-RCNN \cite{ren2015faster} & 27.9& 27.5& 43.1& 16.6& 15.1&5.6 &21.0 &19.6\\
\multirow{8}{*}{}R-CLIP \cite{zhong2022regionclip}& 40.6 \tiny{\textcolor{blue}{(+12.7)}}& 31.3 \tiny{\textcolor{blue}{(+3.8)}}& 47.9 \tiny{\textcolor{blue}{(+4.8)}}& 16.8 \tiny{\textcolor{blue}{(+0.2)}}& 12.0 \tiny{\textcolor{red}{(-3.1)}} &11.2 \tiny{\textcolor{blue}{(+5.6)}} &23.2 \tiny{\textcolor{blue}{(+2.2)}} &26.1 \tiny{\textcolor{blue}{(+6.5)}} \\
\hline
 DIDN \cite{lin2021domain}  \tiny(ICCV'21)  &34.5 \tiny{\textcolor{blue}{(+6.6)}}&30.4 \tiny{\textcolor{blue}{(+2.9)}}&44.2 \tiny{\textcolor{blue}{(+1.1)}} &\textbf{21.2} \tiny{\textcolor{blue}{(+4.6)}}&\textbf{19.0} \tiny{\textcolor{blue}{(+3.9)}}&9.2 \tiny{\textcolor{blue}{(+3.6)}}&22.8 \tiny{\textcolor{blue}{(+1.8)}} & 22.7 \tiny{\textcolor{blue}{(+3.1)}}\\

\textbf{Ours}  & \textbf{41.4} \tiny{\textcolor{blue}{(+13.5)}}& \textbf{31.7} \tiny{\textcolor{blue}{(+4.2)}}& \textbf{49.8} \tiny{\textcolor{blue}{(+6.7)}}& 18.1 \tiny{\textcolor{blue}{(+1.5)}}& 11.4 \tiny{\textcolor{red}{(-3.7)}}& \textbf{12.4} \tiny{\textcolor{blue}{(+6.8)}}& \textbf{25.6} \tiny{\textcolor{blue}{(+4.6)}}&\textbf{27.1} \tiny{\textcolor{blue}{(+7.5)}}\\
\hline
\end{tabular}
\end{table}

\textbf{Domain generalization.} 
 Table~\ref{tab:DG-voc} showcases the generalizability of our proposed method on real-to-artistic tasks. The table is divided into three sections according to the unlabeled domain (i.e., Clipart, Watercolor, and Comic). Our proposed method consistently enhances R-CLIP generalizability and achieves the best result compared to the DA baselines by a large margin. For example, it achieves an $11.7\%$ and $9.3\%$ improvement on Comic when Clipart and Watercolor is the unlabeled source domain, respectively. We also observe that baselines perform better on Watercolor, which is intuitive as Watercolor is the closest domain to VOC. Likewise, our proposed approach outperforms DIDN on \emph{City, Foggy $\rightarrow$ Bdd} by $4.4\%$ (Table~\ref{tab:DG-city}). These results demonstrate the effectiveness of language-guided feature alignment and the proposed consistency learning method for developing a generalizable object detector.

\begin{table}[!tp]\centering
\scriptsize
\captionsetup{skip=0pt, position=above} 
\caption{\textbf{Adverse weather adaptation.}\emph{ City\textrightarrow Foggy} in mAP ($\%$), $\dagger$ from \cite{he2022cross}.}
\label{tab:uda-c2fc}
\begin{tabular}{l|@{\hspace{3pt}}l@{\hspace{2pt}}|cccccccc|l}
\hline
 &{Method}  &{prsn} &{rider} &{car} &{truck} &{bus} &{train} &{motor} 
&{bike} & {mAP} \\
\hline
\multirow{2}{*}{-}&F-RCNN$\dagger$ \cite{ren2015faster}& 36.9 & 36.1 & 44.5.6& 21.7& 32.3&9.2 & 21.5& 32.4&28.3 \textcolor{red}{\tiny{(-20.8)}} \\
&R-CLIP \cite{zhong2022regionclip} &46.5 &51.8 &57.6 &27.3 &45.1 & 19.7& 34.8& 50.2 &\textbf{41.6} \textcolor{red}{\tiny{(-7.5)}} \\
\hline
 \multirow{9}{*}{DA}&SW-DA \cite{saito2019strong} \tiny(CVPR'19) &31.8 & 44.3&48.9 &21.0 &43.8 &28.0&28.9 &35.8& 35.3\\
 &D\&Match$\dagger$ \cite{kim2019diversify} \tiny(CVPR'19)  &31.8 &40.5 &51.0 &20.9  &41.8 &34.3 &26.6 &32.4 & 34.9\\
 &SC-DA$\dagger$ \cite{zhu2019adapting} \tiny(CVPR'19)&33.8&42.1&52.1&26.8&42.5&26.5&29.2&34.4&35.9\\
 &MTOR \cite{cai2019exploring} \tiny(CVPR'19)  &30.6 &41.4 &44.0 &21.9  &38.6 &40.6 &28.3 &35.6 & 35.1\\
 &AFAN$\dagger$ \cite{wang2021afan} \tiny{(TIP'21)}&42.5&44.6&57.0&26.4&48.0&28.3&33.2&37.1&39.6\\
&GPA$\dagger$ \cite{xu2020cross} \tiny(CVPR'20)  & 32.9&46.7 &54.1 &24.7 &45.7 &41.1 &32.4 &38.7 & 39.5\\ 
 &ViSGA$\dagger$ \cite{Rezaeianaran_2021_ICCV} \tiny{(ICCV'21)}&38.8&45.9&57.2&29.9&50.2&\textbf{51.9}&31.9&40.9&43.3\\
 &SFA$\dagger$ \cite{wang2021exploring} \tiny{(acmmm'21)}&46.5&48.6&62.6&25.1&46.2&29.4&28.3&44.0&41.3\\
&DSS$\dagger $\cite{wang2021domain} \tiny{(CVPR'21)}&\textbf{50.9}&\textbf{57.6}&61.1&\textbf{35.4}&50.9&36.6&38.4&51.1&47.8\\
 
 
 &TTD+FPN$\dagger$ \cite{he2022cross} \tiny(CVPR'22) & 50.7 & 53.7 & \textbf{68.2} & 35.1 & 53.0 & 45.1 & 38.9 &49.1 & \textbf{49.2}\\
 &IRG\tablefootnote{Source-free domain adaptation} \cite{vs2022instance} \tiny(CVPR'23)  & 37.4& 45.2& 51.9& 24.4& 39.6& 25.2& 31.5 &41.6 &  37.1\\
 \hline
 \multirow{2}{*}{DG}&DIDN \cite{lin2021domain} \tiny(ICCV'21)  &38.3 &44.4 &51.8 &28.7 &53.3 &34.7 &32.4 & 40.4&40.5 \textcolor{red}{\tiny{(-8.6)}} \\
&\textbf{Ours}  &50.5 &55.1 &66.9 &35.0 &\textbf{56.2} &33.5 & \textbf{41.0}& \textbf{54.3}&\textbf{49.1} \\
\hline
\end{tabular}
\end{table}
\textbf{Extension to domain adaptation.} Our method can naturally extend to domain adaptation as it requires only one labeled source domain. The labeled and unlabeled source domains in the domain generalization task serve as source and target domains, respectively. Table~\ref{tab:uda-c2fc} presents the result of \emph{$City \rightarrow Foggy$} adaptation against baselines and sota DA methods, while real-to-artistic adaptation can be found in the Table~\ref{tab:DA-voc}. As illustrated in Table~\ref{tab:uda-c2fc}, our method improves F-RCNN and R-CLIP by $20.8\%$ and $7.5\%$.
It not only substantially outperforms DIDN (sota DGOD on this setting) by $8.6\%$ but also achieves comparable results to the existing DA works. While CDDMSL is slightly under-performed compared to TTD by only $0.1\%$, contrary to DA methods, it is not designed to adapt to a particular domain.

\subsection{Ablation \& Discussion}
\label{sec:ablation}
\begin{table*}[!tp]
\scriptsize
\captionsetup{skip=0pt, position=above} 
\caption{(a) \textbf{Real-to-artistic adaptation}. VOC is labeled source domain. (b) \textbf{Ablation study}. \emph{VOC, Clipart~$\rightarrow$~Watercolor, Comic (DG), Clipart (DA)}. Results in mAP (\%).}
\begin{subtable}[b]{0.455\linewidth}
\captionsetup{skip=0pt, position=above} 

\begin{tabular}
{|l@{\hspace{3pt}}|c|c@{\hspace{3pt}}|c@{\hspace{0pt}}|} 
\hline
\multirow{2}{*}{Method}& \multicolumn{3}{c|}{Target Domain}\\

 &  \multicolumn{1}{c|}{{Clipart}} & \multicolumn{1}{c|}{{Watercolor}} &\multicolumn{1}{c|}{{Comic}}\\
\hline

F-RCNN \cite{ren2015faster}  &24.1&41.2&17.9 \\
R-CLIP \cite{zhong2022regionclip} &  33.3&44.7&34.2 \\
\hline
Adaptive MT \cite{li2022cross} &30.5&43.7&23.4\\
IRG \cite{vs2022instance}&31.5&\textbf{53.0}&- \\
\hline
 DVA &36.6&43.9&35.9\\
 Caption-PL &35.2&44.2&34.2\\
    \bfseries Ours & \bfseries40.4&49.7&\bfseries46.3\\

\hline
 \end{tabular}
 \caption{Domain adaptation on real-to-artistic}
    \label{tab:DA-voc}
\end{subtable}
\hfill
\begin{subtable}[b]{0.56\linewidth}
\centering
    \begin{tabular}{|@{\hspace{2pt}}l@{\hspace{3pt}}l@{\hspace{3pt}}l@{\hspace{3pt}}l@{\hspace{3pt}}|@{\hspace{3pt}}c@{\hspace{3pt}}|c@{\hspace{3pt}}c@{\hspace{3pt}}|}
        \hline
        &&&  &DA & \multicolumn{2}{c|}{DG}\\
        R-CLIP init. & $\mathcal{L}_{img-cont}$ &  $\mathcal{L}_{inst-cont}$ & $\mathcal{L}_{dist}$ & Clipart & Watercolor&Comic \\
        \hline
        &&&  &24.1&41.2&17.9\\
        $\checkmark$&&&& 32.3&44.7&34.2\\
        \hline
        $\checkmark$& $\checkmark$&&& 32.3&41.7&35.1 \\
        $\checkmark$& &$\checkmark$& & 34.6&45.0&35.4 \\
        $\checkmark$& $\checkmark$&$\checkmark$&&35.1&44.2&35.7 \\
        $\checkmark$& $\checkmark$&$\checkmark$ &$\checkmark$& \textbf{40.4}&\textbf{49.8}&\textbf{45.9} \\
\hline
\end{tabular}
        \caption{Effectiveness of each component}
        \label{tab:ablation_b}
    \end{subtable}
    \vspace{-10.0pt}
\end{table*}

We conduct an extensive ablation study by examining five aspects of CDDMSL: (i) Impact of vision-language pre-training (Sec.~\ref{sec:supervised}), (ii) Effectiveness of language space alignment compared to visual-space alignment (Sec.~\ref{sec:CDDMSL}), (iii) Benefit of feature-space objective compared to token-space objective, (iv) Effectiveness of multi-scale learning (Sec~\ref{sec:CDDMSL}), (v) Contribution of KD regularization (Sec.~\ref{sec:KD}). Table~\ref{tab:ablation_b} exhibits the effectiveness of each component.

(i) \textbf{Vision-language pre-training.} According to Table~\ref{tab:DG-voc}, R-CLIP solely notably improves the generalizability of F-RCNN (pre-trained on Imagenet) by $9.8\%$, $3.5\%$, and $16.3\%$ on Clipart, Watercolor, and Comic, respectively. Additionally, we observe $6.5\%$ gain on \emph{City,Foggy $\rightarrow$ Bdd} (Table~\ref{tab:DG-city}), supporting the benefit of vision-language pre-training. 
Despite the improvement, vision-language pre-training does not guarantee optimal generalizability as it lacks explicit optimization for adaptation to a new domain, and the model may still overfit to domain-specific features during fine-tuning. Hence, we further improve R-CLIP initialization by incorporating descriptive consistency learning as described in Sec.~\ref{sec:CDDMSL}.

(ii) \textbf{Language space vs. visual space alignment.}
    To better understand the effectiveness of enforcing domain-invariant learning through language space, we compare our model with the Direct Visual Alignment baseline (row 5 in Table~\ref{tab:DG-voc} and Table~\ref{tab:DA-voc}) under an identical setting. In all settings, our (last row) substantially exceeds DVA in DA and DG tasks. 

(iii) \textbf{Feature-space vs. token-space.} Comparing Caption-PL (row 4) with our approach in Table~\ref{tab:DG-voc} and Table~\ref{tab:DA-voc} indicates the superiority of enforcing consistency in the feature space. Ours significantly improves Caption-PL in all settings and tasks. Token-space approaches may degrade the performance by focusing on unnecessary non-informative tokens. 

(iv) \textbf{Effectiveness of Multi-Scale Learning.} Table~\ref{tab:ablation_b} shows instance-level is slightly more effective than image-level, but combined delivers the best result. This underscores the importance of mitigating bias and distribution shifts at both levels in object detection.

(v) \textbf{KD regularization's contribution.} Finally, adding a KD regularizer results in the best performance (compare last two rows in Table~\ref{tab:ablation_b}), implying that KD regularization helps the model avoid trivial solutions and produce semantically consistent descriptions.


\begin{table}[!tp]
\scriptsize
\centering
\captionsetup{skip=0pt, position=above} 
\caption{\textbf{Visualization.} \emph{City, Foggy$\rightarrow$ Bdd} \& \emph{VOC, Clipart$\rightarrow$ Comic, Watercolor}.}
\label{Fig:Vis_weather}
\begin{tabular}{@{\hspace{0pt}}c@{\hspace{3pt}}c@{\hspace{2pt}}c@{\hspace{2pt}}}


\raisebox{5.5\height}{\textbf{Night}}& \includegraphics[width=0.32\textwidth]{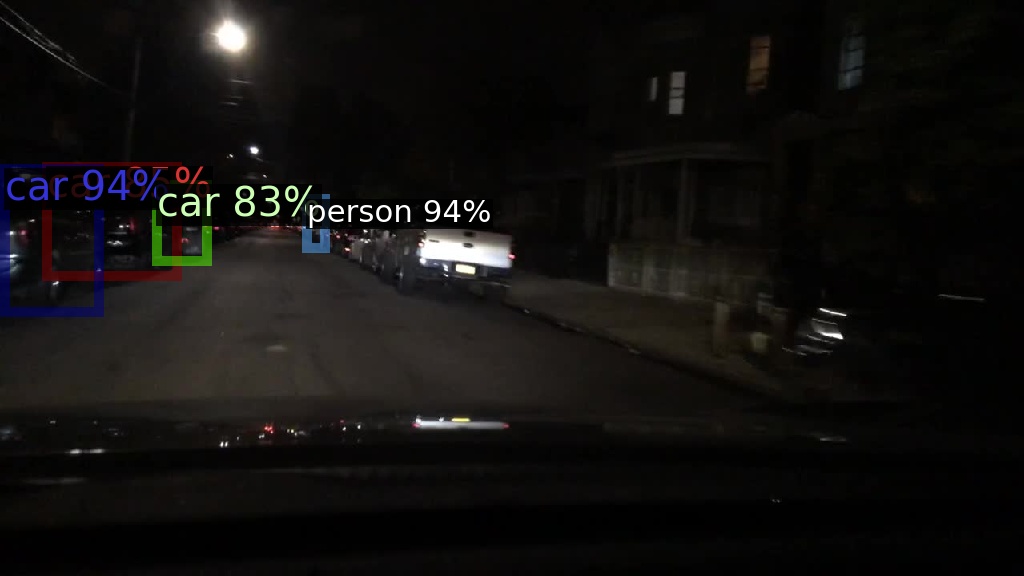} &
\includegraphics[width=0.32\textwidth,]{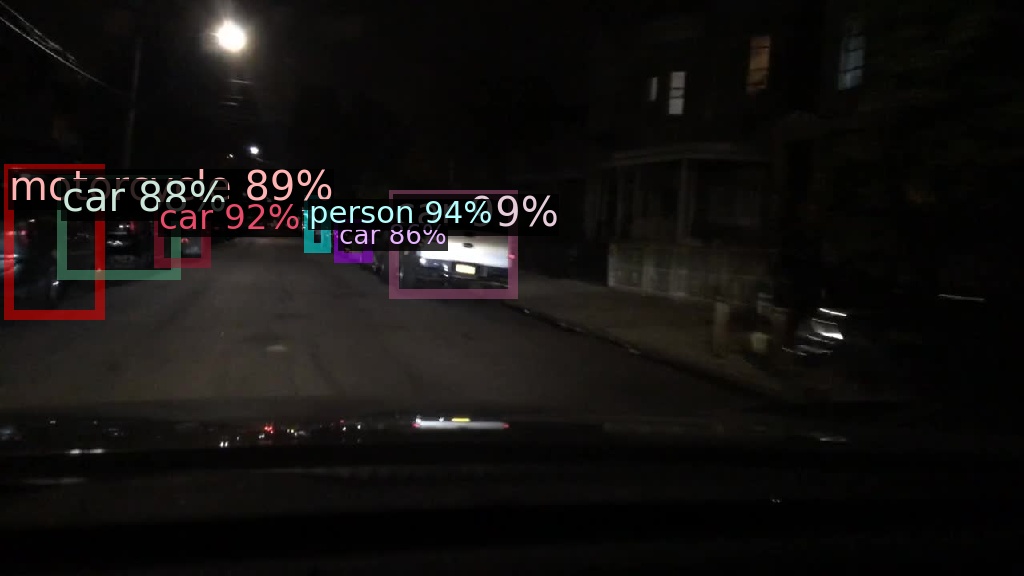} \\
\raisebox{5.5\height}{\textbf{Rainy}}& \includegraphics[width=0.32\textwidth]{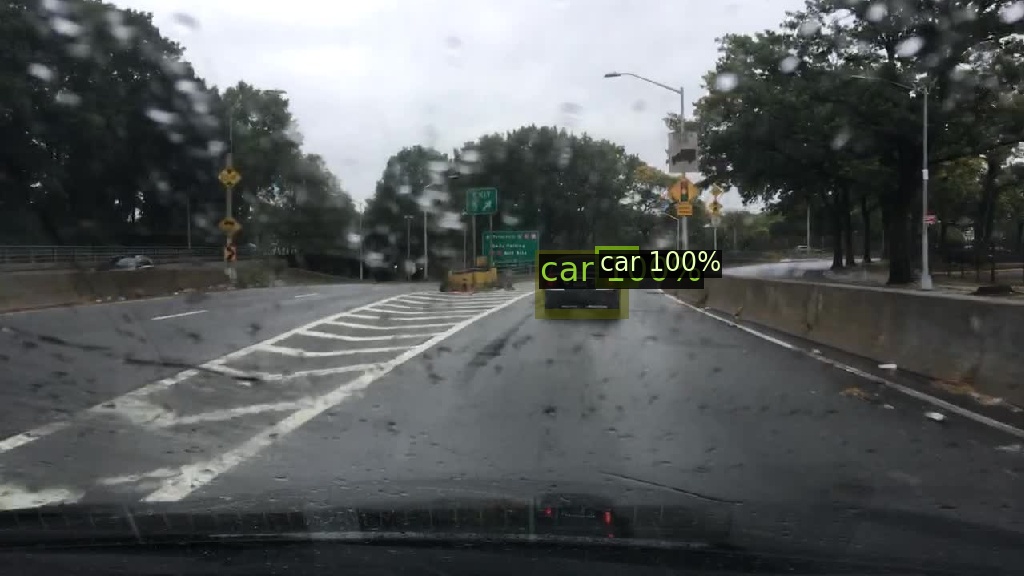} &
\includegraphics[width=0.32\textwidth]{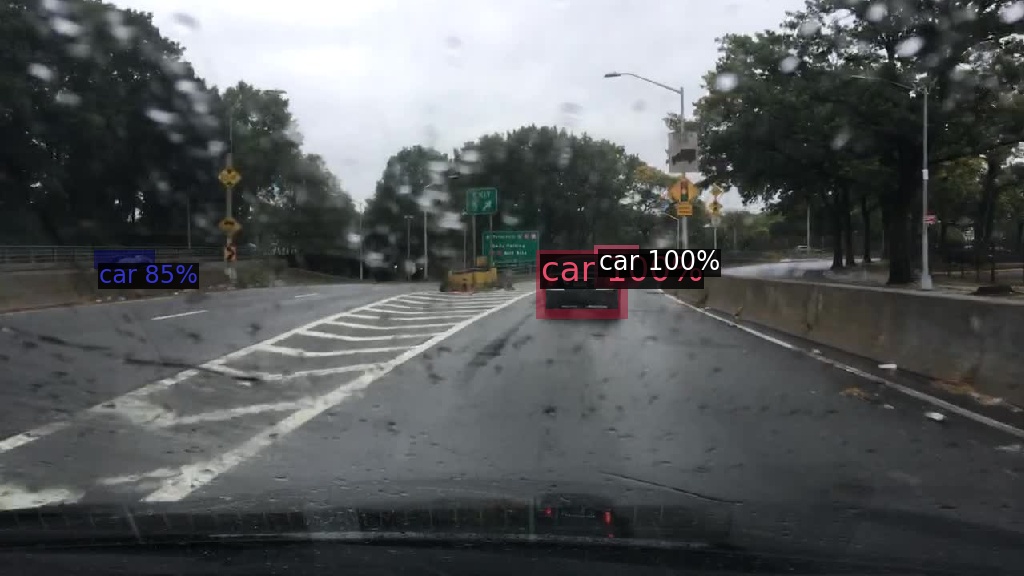} \\
\raisebox{5.5\height}{\textbf{Comic}} & \includegraphics[width=0.29\textwidth,height=0.175\textwidth]{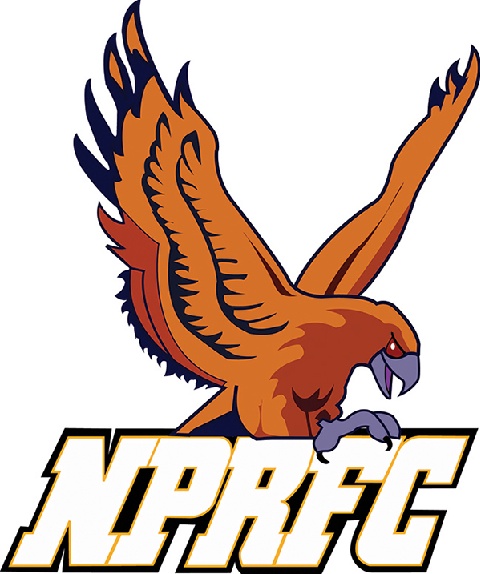} & 
\includegraphics[width=0.29\textwidth,height=0.175\textwidth]{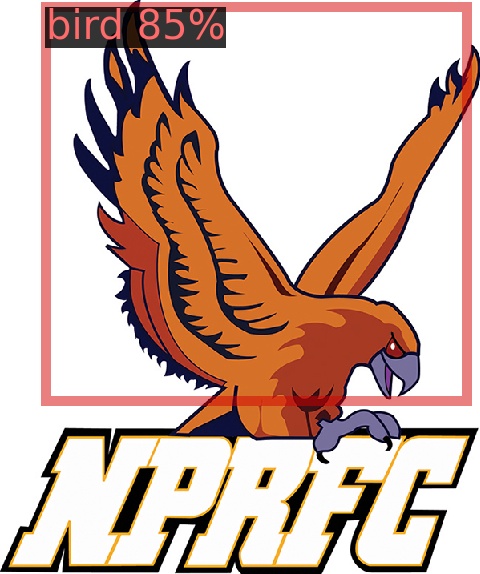}  \\
\raisebox{5.5\height}{\textbf{Watercolor}}& \includegraphics[width=0.32\textwidth,height=0.18\textwidth]{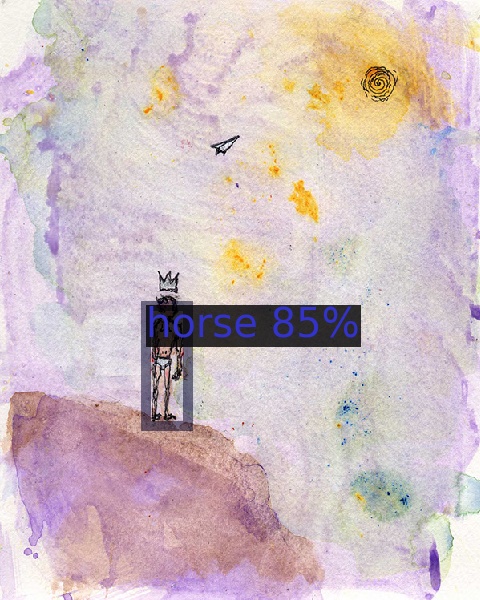} &
\includegraphics[width=0.32\textwidth,height=0.18\textwidth]{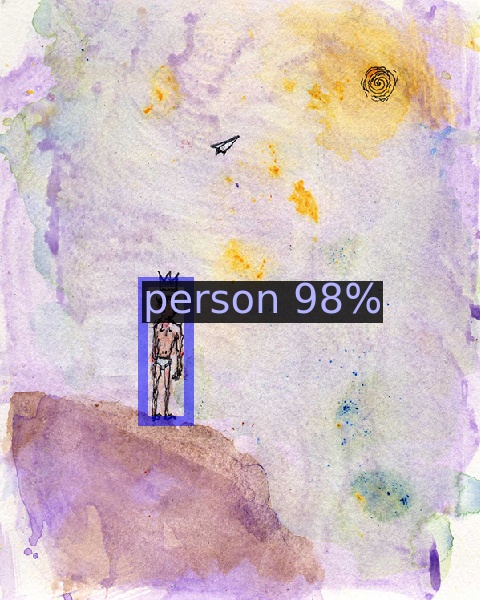}\\
& \textbf{R-CLIP} & \textbf{Ours}\\
\end{tabular}
\end{table}

\textbf{Visualization.} Table~\ref{Fig:Vis_weather} shows qualitative detection results from the R-CLIP baseline and CDDMSL. R-CLIP yields more false negatives, particularly in challenging conditions like rain and detecting distant small objects. Our multi-scale feature alignment improves object-background differentiation, leading to superior detection on the target domain.    Additionally, CDDMSL ensures more accurate recognition by retaining semantically crucial information.
\section{Conclusion}
This paper introduces a novel approach, termed Cross-Domain Descriptive Multi-Scale Learning (CDDMSL), for semi-supervised domain generalization in object detection. Initially, we explore the effectiveness of vision-language pre-training in achieving a robust object detector. Subsequently, we present a unique method that promotes domain-invariant feature learning on the visual encoder through the language space. This is achieved using a contrastive learning-based approach at both image and instance levels. CDDMSL delivers state-of-the-art results in DA and DG tasks across various object detection benchmarks.

\noindent \textbf{Acknowledgement:} The work is in part supported by NSF Grant No. 2006885.


\bibliography{mybib}



\end{document}


\maketitle







\section{Baselines}
We evaluate our proposed model against source-only fully-supervised object detection models (FSOD), domain adaptation (DA), and domain generalization (DG) baselines on two benchmarks: \textit{real-to-artistic} and \textit{adverse weather}. This project leverages a ResNet50 backbone pre-trained by RegionCLIP; therefore, for a fair comparison, we only compare it with related works utilizing the ResNet50 backbone.  Notably, larger ResNet models like ResNet101 are not yet provided for RegionCLIP \cite{zhong2022regionclip}.
To our knowledge, DIDN \cite{lin2021domain} is the only multi-domain DG work in object detection. However, DIDN only conducted experiments on adverse weather, and its implementation code is not publicly available, preventing fair comparison on \textit{real-to-artistic}. Thus, on the DG task in real-to-artistic transfer, we compare to source-only FSOD baselines (i.e., Faster-RCNN and RegionCLIP), domain adaptation methods, namely Adaptive-MT \cite{he2022cross}, IRG \cite{he2022cross}, and our defined DG baselines, namely \textit{Direct Visual Alignment (DVA)} and \textit{Caption Pseudo-Labeling (Caption-PL)}. 

Extending our method to DA, we compare it across the \textit{real-to-artistic} domain with FSOD baselines, DA methods, and our DG benchmarks. We further compare our model on \textit{adverse weather} DG task with source-only FSOD baselines and DIDN \cite{lin2021domain}. Following \cite{lin2021domain}, and since most of the previous works have used these benchmarks in their experiments, we have extensively compared our model on the \textit{City $\rightarrow$ Foggy} adaptation task, against both source-only FSOD and DA baselines. SW-DA \cite{saito2019strong}, D\&Match \cite{kim2019diversify}, MTOR \cite{cai2019exploring}, AFAN \cite{wang2021afan}, GPA \cite{xu2020cross}, SFA \cite{wang2021exploring}, DSS \cite{wang2021domain}, TTD+FPN \cite{he2022cross}, and IRG \cite{he2022cross} are the DA state-of-the-arts that are compared with on this task.


\section{Qualitative results}

\subsection{Stability comparison}
\begin{figure}
    \centering
    \includegraphics[width=0.6\textwidth]{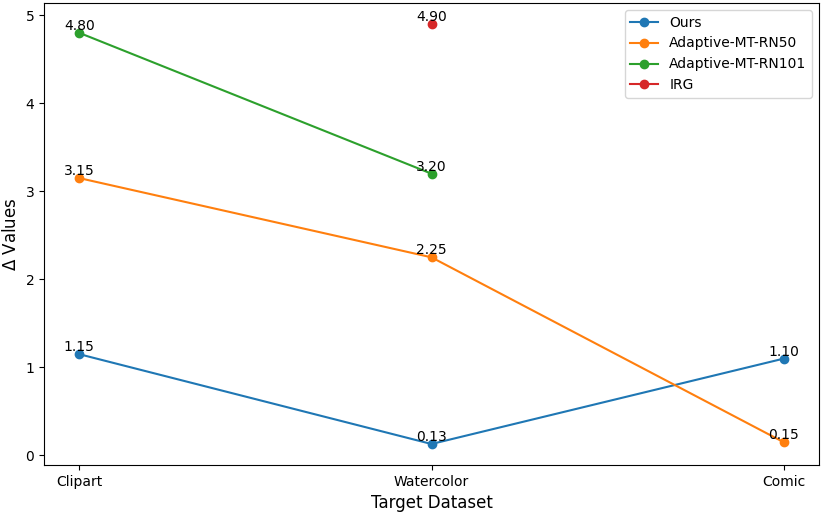} 
    \caption{\textbf{Comparison of $\Delta$}. $\Delta$ represents the drop in the performance on the target domain when the target is used as the source domain (i.e., DA) and when it is unknown throughout the training (i.e., DG). Hence, a lower number means that the model is more stable. Adaptive-MT-RN101 and IRG values are borrowed from \cite{li2022cross} and \cite{vs2022instance}, respectively. Since values for all settings were not reported, $\Delta$ is reported for Clipart and Watercolor for Adaptive-MT-RN101 and Watercolor for IRG. RN101 is the ResNet101 backbone, while RN50 is the ResNet50 backbone.}
    \label{fig:delta_comparison}
\end{figure}

Fig.~\ref{fig:delta_comparison} demonstrates the stability of our method compared to the domain adaptation methods on the real-to-artistic benchmark. We define $\Delta = DA_{mAP} - DG_{mAP}$. For example, for Clipart, we subtract the average performance on Clipart when Clipart is used as a target domain in the DG task (i.e., Table~1 in the main text) from the performance on Clipart in a DA task (i.e., Table~4.a in the main text). Our model mAP drop on DG sets is relatively much lower than the DA methods, especially on Clipart and Watercolor. For example, on Clipart, our model's performance drops by 1.15\%, while Adaptive-MT-RN50's performance drops by 3.15\% and Adaptive-MT-RN101's performance drops by 4.8\%. Comic $\Delta$ for our model is higher than Adaptive-MT-RN50; this is due to the fact that Adaptive-MT-RN50 performs extremely poorly on Comic in both DA and DG settings. For instance, in the DA task, our model achieves 46.3\% while Adaptive-MT-RN50 achieves 23.4\% (Table~4.a), and the average performance in the DG settings is 45.2\% and 23.25\% for ours and Adaptive-MT-RN50, respectively (Table~1 in the main text).

\subsection{Caption comparison}
Table~\ref{tab:caption-compare} includes some examples of the generated caption on the validation set of VOC, Clipart, Watercolor, and Comic. Specifically, we use ClipCap to generate captions based on RegionCLIP baseline trained on VOC in a supervised manner and when our method is trained on \textit{VOC, Clipart $\rightarrow$ Watercolor, Comic}. Even though our method does not produce perfect descriptions, especially in artistic domains, it constantly produces better and more meaningful descriptions than the RegionCLIP baseline, as captions may become meaningless on even natural images from the VOC dataset when FSOD finetuning. For instance, in Table~\ref{tab:caption-compare}, the RegionCLIP baseline generated an unrelated caption for the image from the VOC dataset that is missing the essential information such as "person" and  "horse." This shows the importance of our proposed knowledge distillation-based regularization technique to ensure that captions are meaningful and related to the corresponding image.  

\begin{table}[!tp]
\scriptsize
\centering
\captionsetup{skip=0pt, position=above} 
\caption{\textbf{Caption Comparison}. Comparing generated caption based on RegionCLIP baseline and CDDMSL visual features when training using VOC and Clipart as source domains }
\label{tab:caption-compare}
\begin{tabular}{@{\hspace{0pt}}l@{\hspace{2pt}}c@{\hspace{0pt}}c@{\hspace{0pt}}c@{\hspace{0pt}}c@{\hspace{0pt}}}
 & VOC &Clipart&Watercolor&Comic\\ 

& \includegraphics[width=0.115\textwidth]{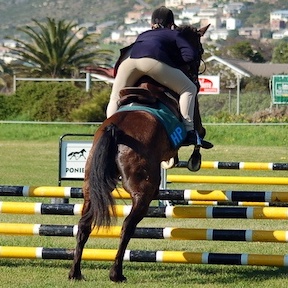} &
\includegraphics[width=0.115\textwidth]{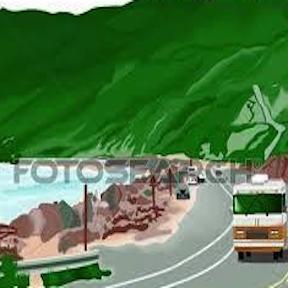} & 
\includegraphics[width=0.12\textwidth]{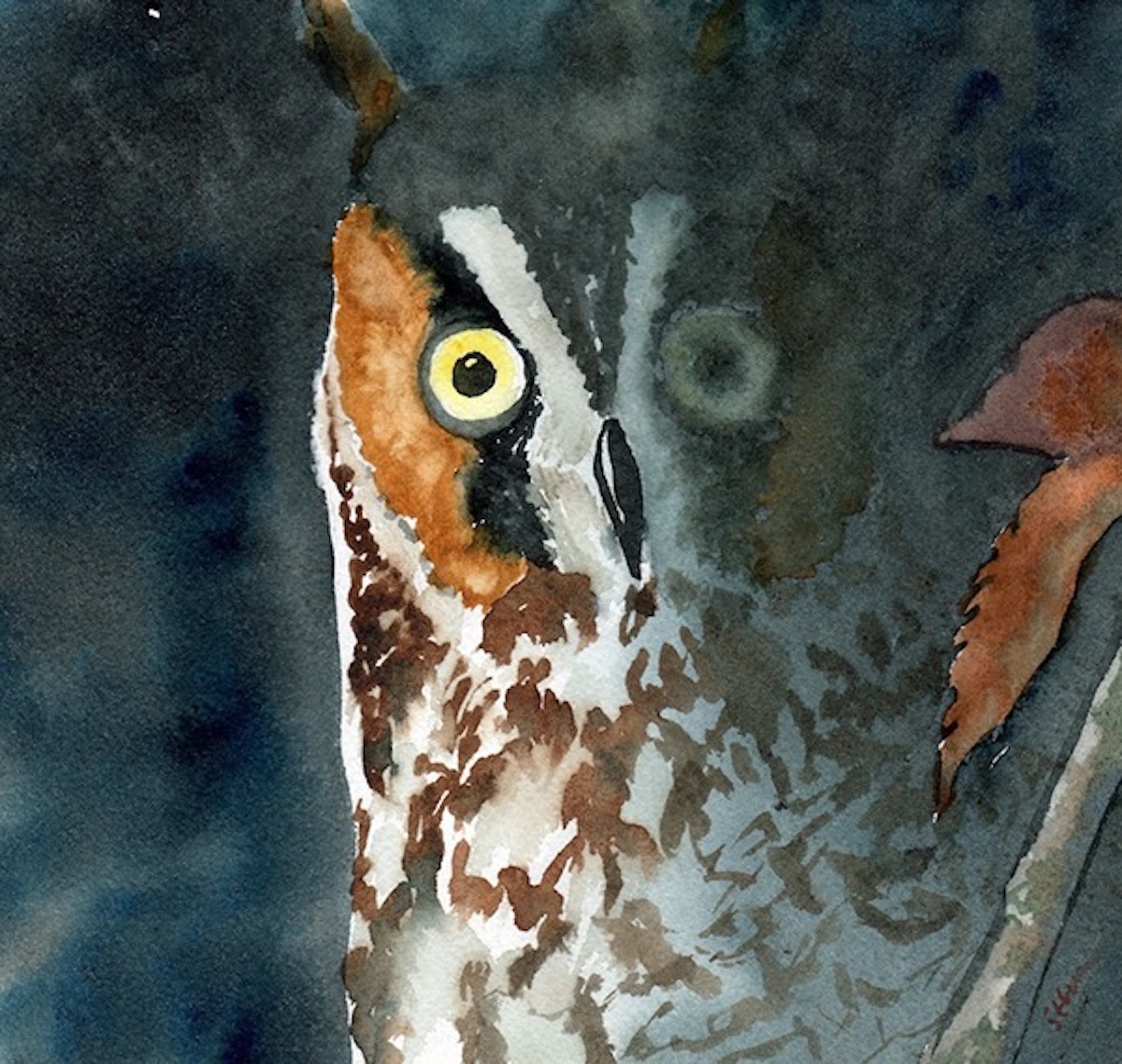} & 
\includegraphics[width=0.115\textwidth]{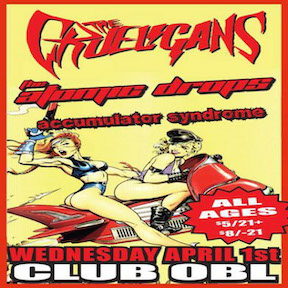} \\

RegionCLIP & \multicolumn{1}{p{1.7cm}}{\raggedright Two people are standing in a room with a table.} & \multicolumn{1}{p{1.7cm}}{\raggedright A row of soccer jerseys on a rooftop} &
\multicolumn{1}{p{1.7cm}}{\raggedright A picture of a wall of knit hats} & \multicolumn{1}{p{1.9cm}}{\raggedright A man is standing in a coffee shop with a group of people watching him.} \\ \hline

Ours &\multicolumn{1}{p{1.7cm}}{\raggedright A person riding a brown horse on a paved field.} &  \multicolumn{1}{p{1.7cm}}{\raggedright A green and white truck driving down a beach.} &
\multicolumn{1}{p{1.7cm}}{\raggedright A brown and white drawing of a bird.} & 
\multicolumn{1}{p{1.7cm}}{\raggedright A man riding a motorcycle on top of a building}\\ \hline
\end{tabular}

\end{table}


    



\begin{table}[!tp]
\scriptsize
\centering
\captionsetup{skip=0pt, position=above} 
\caption{Visualization inference of our model on real-to-artistic. Each pair of columns corresponds to a different sample with RegionCLIP (left column in each pair) trained on labeled data and Ours (right column in each pair) trained on VOC (labeled) and Clipart (unlabeled) for real-to-artistic generalization.}
\label{tab:predictions-compare-voc}

\begin{tabular}{l@{\hspace{2pt}}c@{\hspace{2pt}}c@{\hspace{2pt}}|@{\hspace{2pt}}c@{\hspace{2pt}}c}
\raisebox{1.5cm}{\textbf{Clipart}} &\includegraphics[width=0.2\textwidth,height=0.2\textwidth]{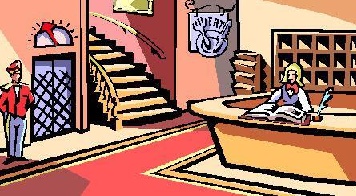} &
\includegraphics[width=0.2\textwidth,height=0.2\textwidth]{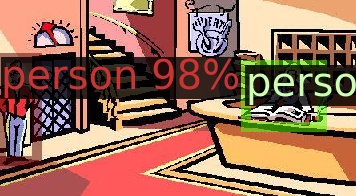} & \includegraphics[width=0.2\textwidth,height=0.2\textwidth]{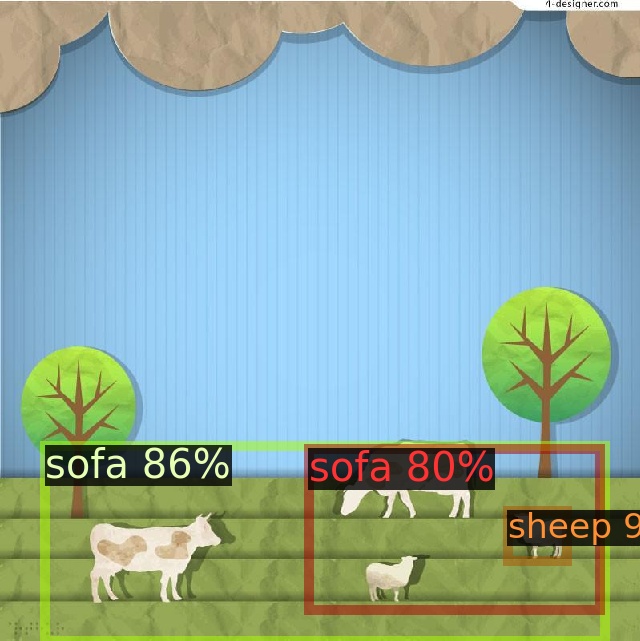} &
\includegraphics[width=0.2\textwidth,height=0.2\textwidth]{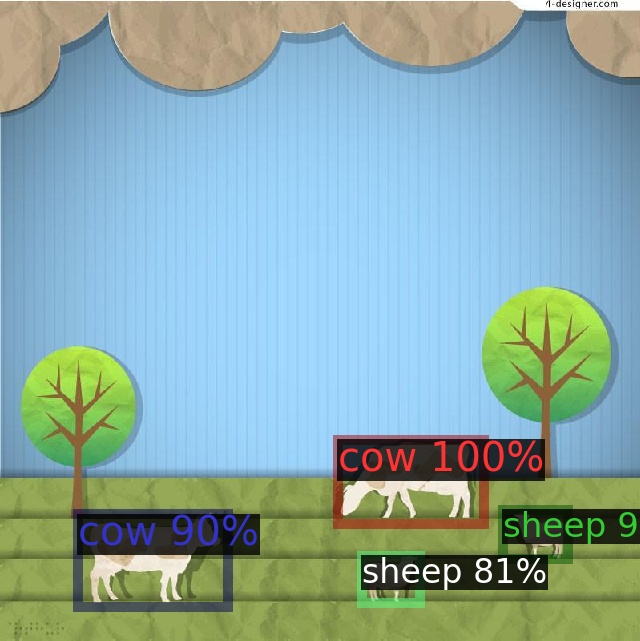} \\

\raisebox{1.5cm}{\textbf{Watercolor}} & \includegraphics[width=0.2\textwidth,height=0.2\textwidth]{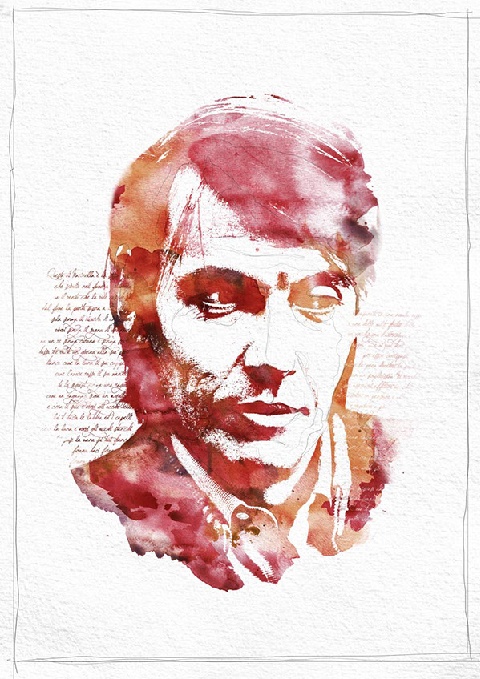} &
\includegraphics[width=0.2\textwidth,height=0.2\textwidth]{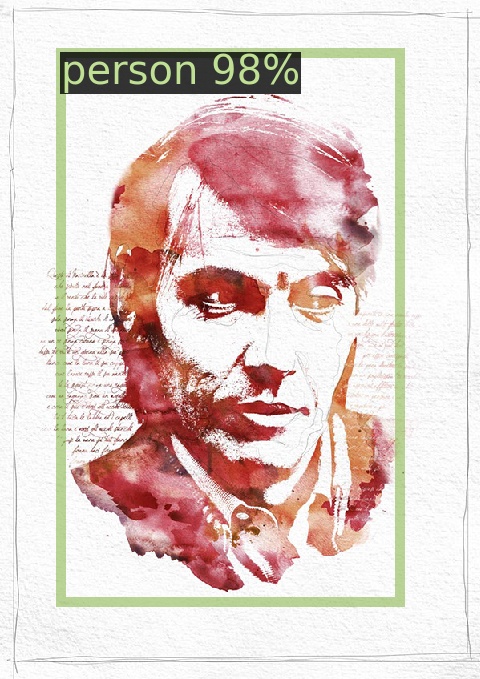}
& \includegraphics[width=0.2\textwidth,height=0.2\textwidth]{images/pred_comp/real-to-artistic/watercolor/baseline/img_155004325.jpg} &
\includegraphics[width=0.2\textwidth,height=0.2\textwidth]{images/pred_comp/real-to-artistic/watercolor/ours/img_155004325.jpg}\\

 \raisebox{1.5cm}{\textbf{Comic}} & \includegraphics[width=0.2\textwidth,height=0.2\textwidth]{images/pred_comp/real-to-artistic/comic/baseline/img_33497847.jpg} & 
\includegraphics[width=0.2\textwidth,height=0.2\textwidth]{images/pred_comp/real-to-artistic/comic/ours/img_33497847.jpg}  
&\includegraphics[width=0.2\textwidth,height=0.2\textwidth]{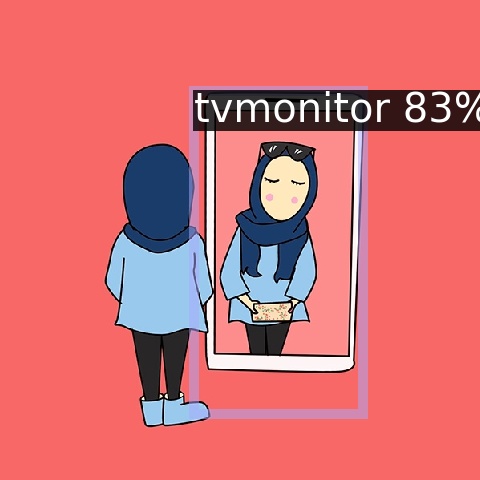} & 
\includegraphics[width=0.2\textwidth,height=0.2\textwidth]{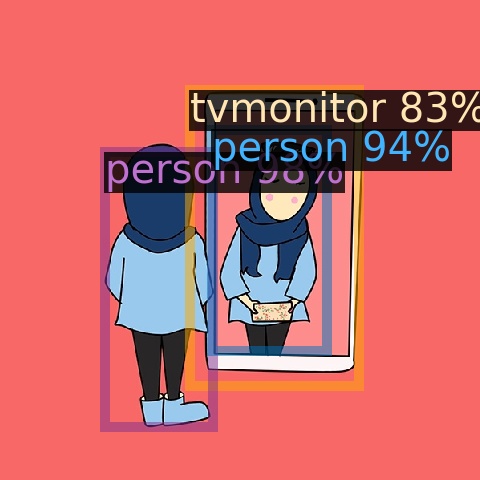}  \\
& \textbf{R-CLIP} & \textbf{Ours}& \textbf{R-CLIP} & \textbf{Ours}\\
\end{tabular}
\end{table}

\begin{table}[!tp]
\scriptsize
\centering
\captionsetup{skip=0pt, position=above} 
\caption{Visualization inference of our model on adverse-weather domains. RegionCLIP (left column) is trained on labeled data. Ours (right column) is trained on City and Foggy for adverse-weather generalization. The first two rows are results from Foggy, and the last three rows are predictions on the Bdd test set.}
\label{tab:predictions-compare-city}
\begin{tabular}{l@{\hspace{2pt}}c@{\hspace{2pt}}c@{\hspace{2pt}}}
& \includegraphics[width=0.5\textwidth]{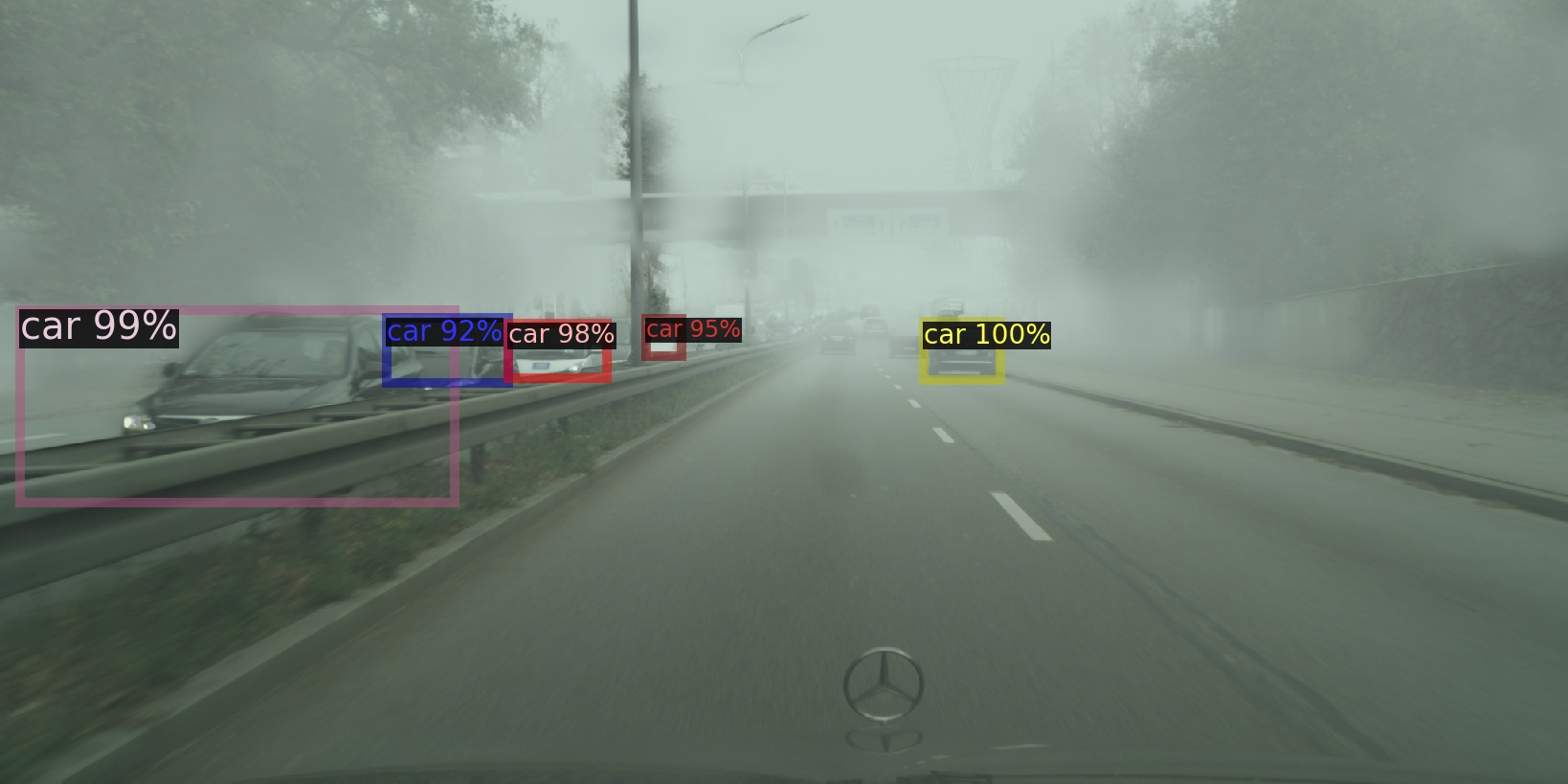} &
\includegraphics[width=0.5\textwidth,]{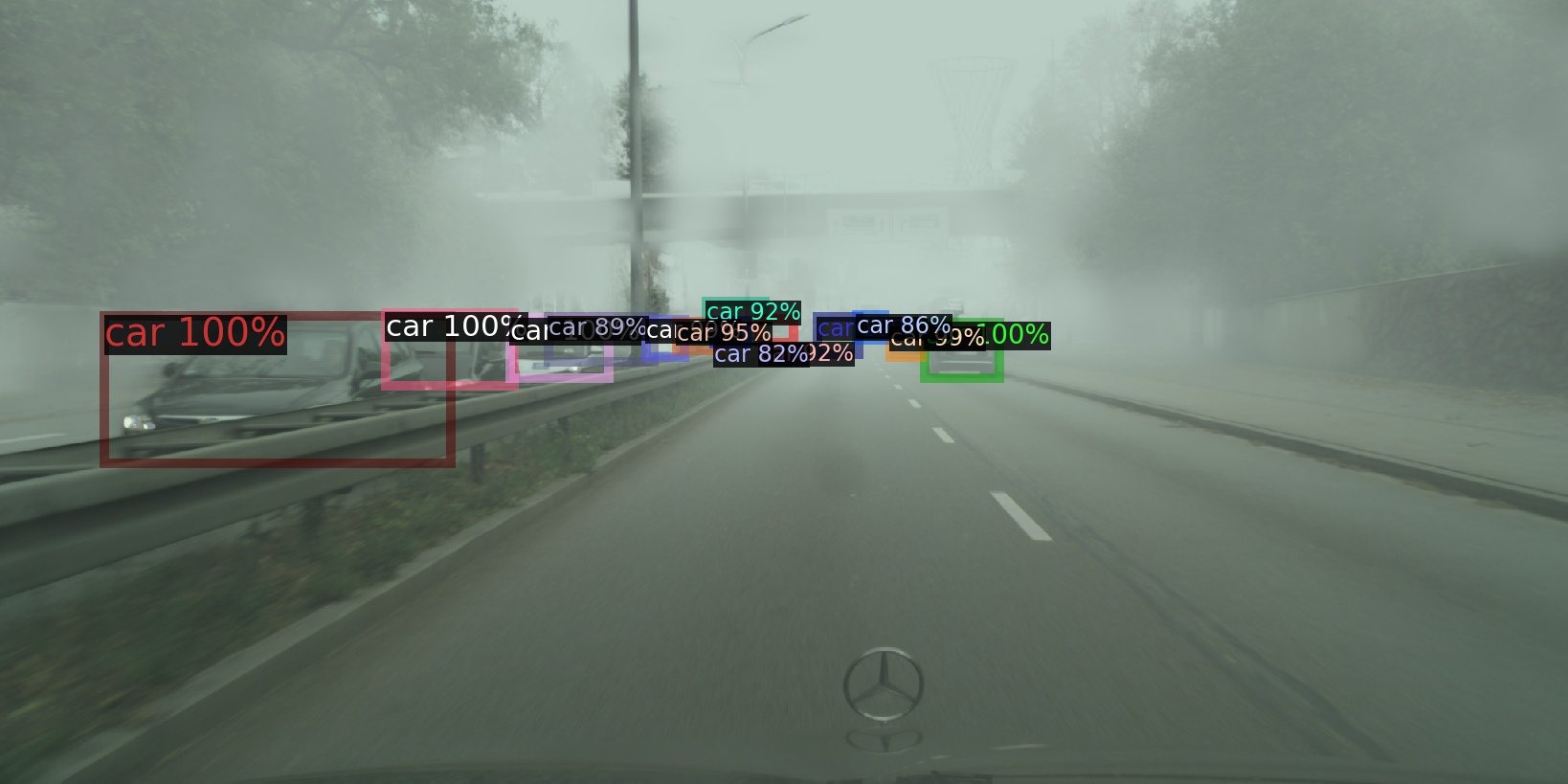} \\

& \includegraphics[width=0.5\textwidth]{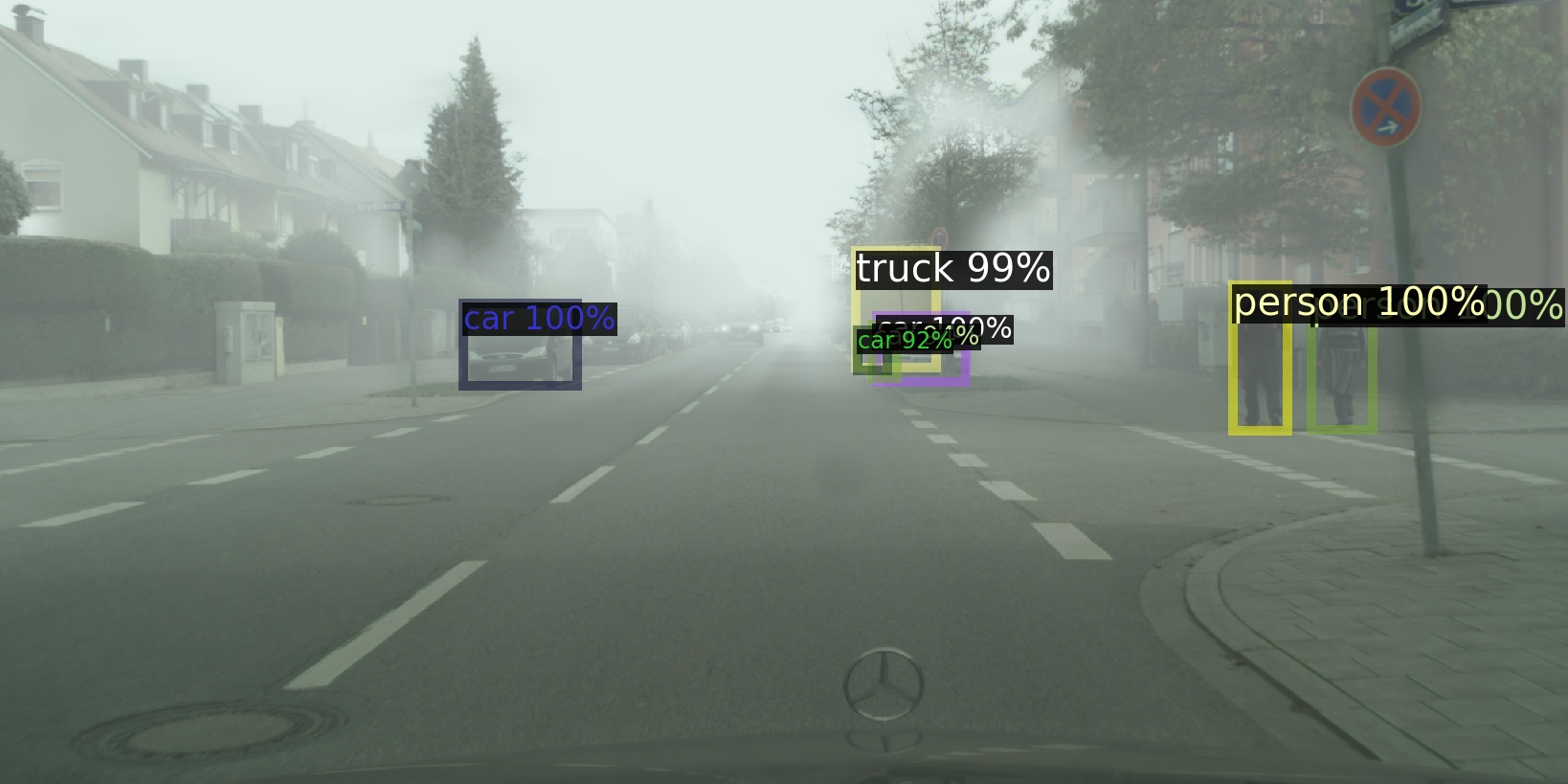} &
\includegraphics[width=0.5\textwidth,]{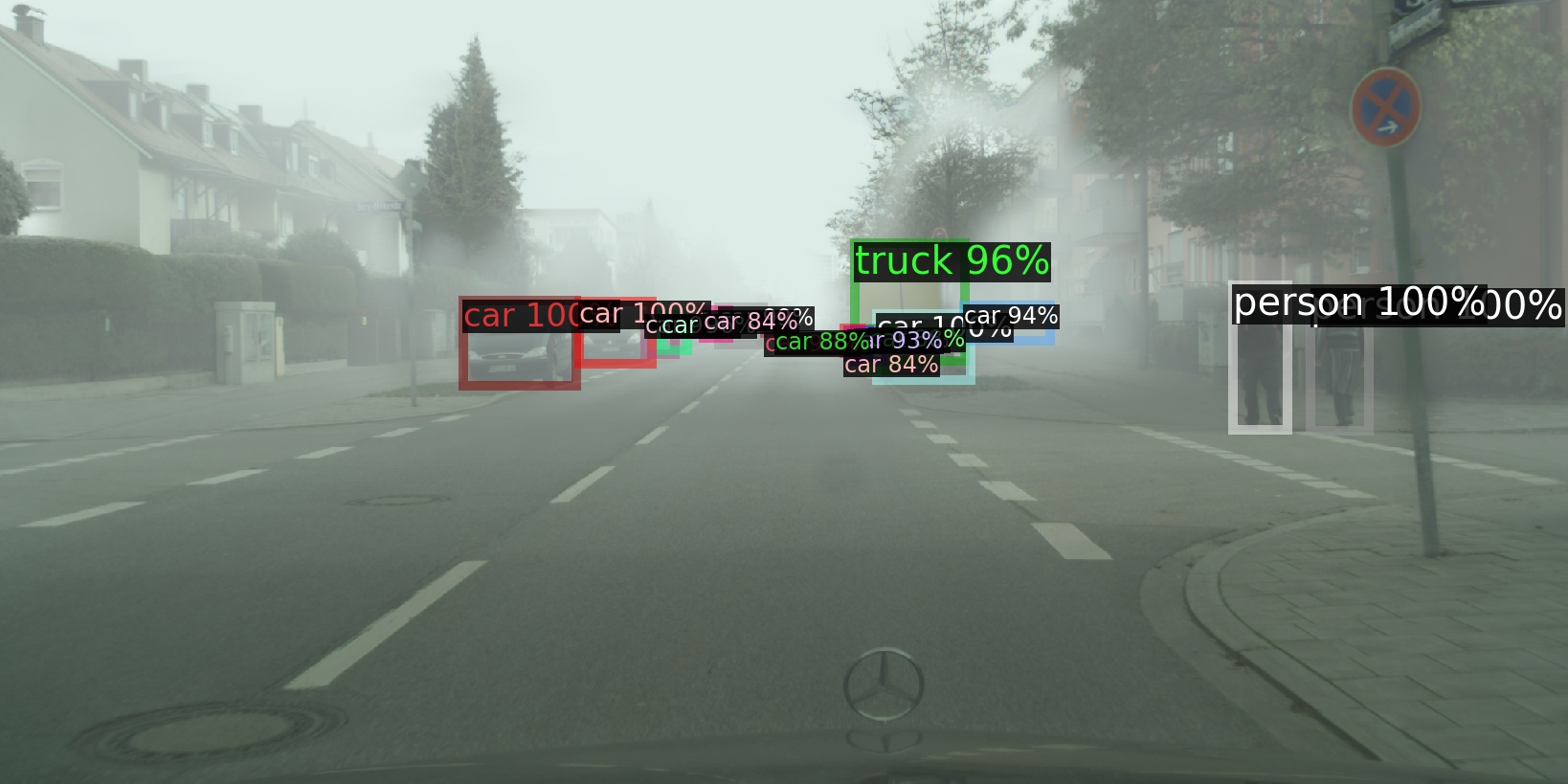} \\

\midrule
& \includegraphics[width=0.5\textwidth]{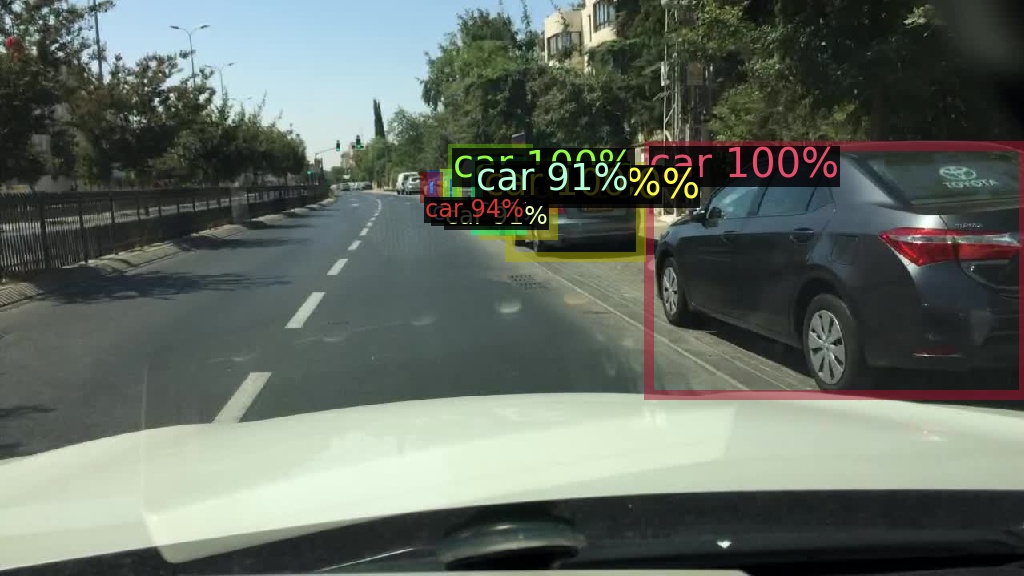} &
\includegraphics[width=0.5\textwidth,]{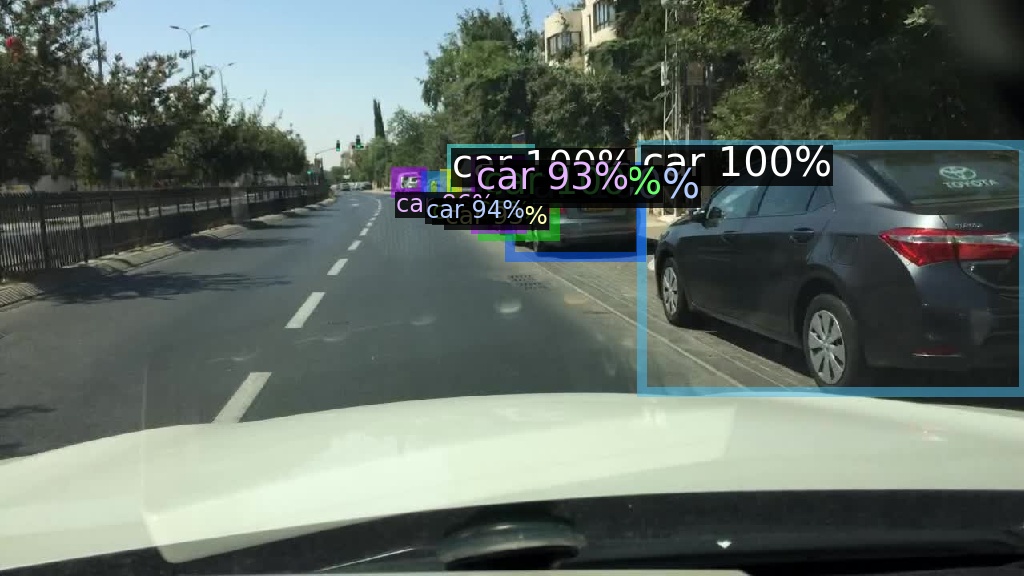} \\

& \includegraphics[width=0.5\textwidth]{images/pred_comp/adverse-weather/bdd/baseline/img_b84f4f83-c6842bd6.jpg.jpg} &
\includegraphics[width=0.5\textwidth,]{images/pred_comp/adverse-weather/bdd/ours/img_b84f4f83-c6842bd6.jpg.jpg} \\

& \includegraphics[width=0.5\textwidth]{images/pred_comp/adverse-weather/bdd/baseline/img_c836f138-d9b4f7a1.jpg.jpg} &
\includegraphics[width=0.5\textwidth,]{images/pred_comp/adverse-weather/bdd/ours/img_c836f138-d9b4f7a1.jpg.jpg} \\

& \textbf{R-CLIP} & \textbf{Ours}\\
\end{tabular}
\end{table}




\clearpage
\bibliography{mybib}